%% file: main.tex
\pdfoutput=1

\documentclass[11pt]{article}

\usepackage[final]{acl}

\usepackage{times}
\usepackage{latexsym}

\usepackage[T1]{fontenc}

\usepackage[utf8]{inputenc}

\usepackage{microtype}

\usepackage{inconsolata}

\usepackage{multirow}
\usepackage{booktabs}
\usepackage[normalem]{ulem}
\useunder{\uline}{\ul}{}
\usepackage{algorithmicx,algorithm}
\usepackage{graphicx}
\usepackage{color}
\usepackage{colortbl}
\usepackage{amsmath}
\usepackage{subcaption}

\usepackage{tikz}
\title{A Survey of Large Language Models in Discipline-specific Research: Challenges, Methods and Opportunities}

\author{
 \textbf{Lu XIANG},
 \textbf{Yang ZHAO},
 \textbf{Yaping ZHANG},
 \textbf{Chengqing ZONG}\Thanks{ Corresponding Author}
\\
 State Key Laboratory of Multimodal Artificial Intelligence Systems, \\ Institute of Automation, Chinese Academy of Sciences, Beijing, China
 \\
\tt {lu.xiang@nlpr.ia.ac.cn}; \tt{\{yang.zhao,yaping.zhang,cqzong\}@nlpr.ia.ac.cn}
}

\input{00_definitions}

\begin{document}
\maketitle
\begin{abstract}
Large Language Models (LLMs) have demonstrated their transformative potential across numerous disciplinary studies, reshaping the existing research methodologies and fostering interdisciplinary collaboration. However, a systematic understanding of their integration into diverse disciplines remains underexplored. This survey paper provides a comprehensive overview of the application of LLMs in interdisciplinary studies, categorising research efforts from both a technical perspective and with regard to their applicability. From a technical standpoint, key methodologies such as supervised fine-tuning, retrieval-augmented generation, agent-based approaches, and tool-use integration are examined, which enhance the adaptability and effectiveness of LLMs in discipline-specific contexts. From the perspective of their applicability, this paper explores how LLMs are contributing to various disciplines including mathematics, physics, chemistry, biology, and the humanities and social sciences, demonstrating their role in discipline-specific tasks. The prevailing challenges are critically examined and the promising research directions are highlighted alongside the recent advances in LLMs. By providing a comprehensive overview of the technical developments and applications in this field, this survey aims to serve as an invaluable resource for the researchers who are navigating the complex landscape of LLMs in the context of interdisciplinary studies.

\end{abstract}

\section{Introduction}

The rapid development of Large Language Models (LLMs) \cite{2023arXiv230308774OpenAI,Touvron2023,zeng2023glmb,yang2024qwen2,bi2024deepseek} has marked a revolutionary leap in the field of artificial intelligence. These models, boasting billions of parameters, have demonstrated a remarkable proficiency in a wide array of AI tasks, including natural language generation, reasoning, image understanding, code generation etc. \cite{Touvron2023,bai2023qwen,li-etal-2024-improving-context,yao2023react}. As a result, LLMs have emerged as the most representative achievement and the core of AI. 

LLMs exhibit several key advantages. First, they excel in language understanding, translation, and generation, enabling them to process and produce human-like language with remarkable fluency. Second, their ability to leverage vast amounts of knowledge allows them to effectively perform general knowledge question answering. Third, LLMs demonstrate strong reasoning capabilities, enabling them to tackle complex problems and provide informed insights. These strengths have driven the increasing application of LLMs in a wide range of scientific and academic disciplines, from mathematics \cite{liu2023mathematical} and physics\cite{Ma2024discovery,xu2025ugphysics} to biology \cite{tinn2023fine,btae075_jin} and the humanities and social sciences \cite{gao2024large,bail2024can}, facilitating new possibilities for interdisciplinary collaboration.

This survey aims to provide a comprehensive overview of LLM technologies and their applications across various disciplines. This study has three primary objectives: (1) to explore the key technical methods used to adapt LLMs to meet the specific needs of various disciplines; (2) to examine practical applications of LLMs in diverse disciplines, including mathematics, physics, chemistry, biology, and the humanities and social sciences, with an emphasis on how LLMs address unique challenges; (3) to outline the challenge and opportunities for interdisciplinary collaboration facilitated by LLMs. 

\begin{table*}[]
\centering
\begin{tabular}{lcc}
\hline
\multicolumn{1}{c}{\textbf{Model Series}}          & \textbf{Developer}  & \textbf{\begin{tabular}[c]{@{}c@{}}Open source \\ Y/N\end{tabular}} \\ \hline
PaLM series \cite{Chowdhery2023} & Google  & N            \\
LLaMA series \cite{Touvron2023,touvron2023llama} & Meta  & Y  \\
GPT-4 \cite{2023arXiv230308774OpenAI}   & OpenAI    & N     \\
GLM-4 \cite{glm2024chatglm}  & Tsinghua University & Y  \\
Qwen series \cite{bai2023qwen,yang2024qwen2} & Alibaba             & Y                        \\
DeepSeek series \cite{liu2024deepseek,liu2024deepseekv3}             & DeepSeek AI         & Y                        \\ \hline
\end{tabular}
\caption{Overview of Common Series-Based LLMs.}
\label{fig:LLMs}
\end{table*}

The remainder of this article is organised as follows. Section \ref{sec2} provides the preliminaries of LLMs and the challenges related to applying LLMs into various disciplines. Following that, the taxonomy of LLM techniques for different disciplines is detailed in Section \ref{sec3}. Section \ref{sec4} introduces application cases across various disciplines, presenting how researchers have practically employed LLMs in Mathematics, Physics, Chemistry, Biology, and the Humanities and Social Sciences. Subsequently, Section \ref{sec5} discusses the critical challenges and outlines future directions for enhancing the effectiveness and applicability of LLMs in interdisciplinary studies. Finally, Section \ref{sec6} concludes this survey paper.

\section{Background} \label{sec2}
\subsection{Preliminaries}

The foundation of most modern LLMs lies in the Transformer model, which was introduced in 2017 \cite{vaswani2017attention,zhao2023transformer} and revolutionized the field of natural language processing. Transformers are capable of efficiently modeling long-range dependencies in various texts, processing sequences in parallel, and improving computational scalability by enabling more efficient training on larger datasets and model sizes. Early transformer-based models like BERT \cite{devlin-etal-2019-bert} and GPT \cite{radford2018improving} established a pre-training and fine-tuning paradigm. 

With the rapid scaling of model parameters, LLMs like GPT-3 \cite{brown2020language}, demonstrated impressive capabilities in zero-shot and few-shot learning. This growth in model size, as illustrated by Scaling Laws \cite{kaplan2020scaling}, has significantly improved the performance of LLMs across various domains, enabling them to tackle increasingly complex tasks. Additionally, techniques like instruction fine-tuning \cite{chung2024scaling} and InstructGPT \cite{Ouyang2022} have further enhanced the model’s task-specific adaptability. The release of ChatGPT \cite{OpenAI2022} in late 2022 marked a key milestone, demonstrating the tremendous potential of LLMs.

\subsection{Popular LLMs}

The continuous advancements in LLMs have been propelled by both academic and industrial research. Table \ref{fig:LLMs} includes some common series-based LLMs. Notably, the release of DeepSeek V3 \cite{liu2024deepseekv3} at the end of 2024 has garnered significant attention due to its innovations related to performance, efficiency, and cost. With a training cost of just \$5.57 million, DeepSeek V3 is recognised as one of the most cost-effective open-source models. It features 671 billion parameters and uses a hybrid expert model (MoE) design, allowing for the efficient processing of complex tasks. DeepSeek V3 excels in areas such as multilingual processing, mathematical problem-solving, and code generation, competing with top closed-source models while maintaining an outstanding performance. 

\subsection{Core Challenges of LLMs for Various Disciplines}

The application of LLMs across different domains or disciplines often encounters several key challenges. One of the most common issues is the lack of domain-specific knowledge. Although LLMs are trained on extensive general-purpose datasets, they often fall short in providing the specialised knowledge necessary for technical fields such as biology, chemistry, or physics \cite{han2024generalist}. This limitation becomes particularly apparent when LLMs attempt to tackle complex problems that require precise and up-to-date information. Moreover, LLMs may generate information that appears to be plausible but is incorrect or fabricated - a phenomenon referred to as "hallucination" \cite{huang2024survey}. This issue is particularly concerning in research contexts, where inaccuracies can lead to significant consequences. Addressing these challenges is crucial for the effective deployment of LLMs in specialised discipline studies.

\input{Figures/category_tree}

\section{Methods of Applying LLMs across Various Disciplines} \label{sec3}

The application of LLMs across various disciplines has been greatly enhanced by different techniques. As shown in Figure \ref{fig:categorization_of_survey}, these techniques can be broadly classified into two categories based on how the LLM interacts with external factors: \textit{Internal Knowledge Optimization} and \textit{External Interaction and Collaboration}. Both categories address domain-specific challenges and enhance LLMs’ performance in specialised tasks.

\subsection{Internal Knowledge Optimisation}

This category focuses on refining and enhancing the inherent knowledge and capabilities of LLMs through techniques like continued pre-training (CPT), supervised fine-tuning (SFT), and reinforcement learning from human feedback (RLHF).

\subsubsection{Continued Pre-training (CPT)}

CPT involves further training a pre-trained LLM on domain-specific data to deepen its expertise in a particular field \cite{jang2022towards}. For example, a general-purpose LLM might be further trained on scientific literature to improve its performance in answering research questions or generating domain-specific content \cite{shao2024deepseekmath, azerbayev2023llemma}. While effective in improving domain-specific performance, CPT is computationally expensive and can lead to catastrophic forgetting challenges.

\subsubsection{Supervised Fine-tuning (SFT)}

SFT adapts pre-trained LLMs to specific tasks or domains by training them on labeled datasets \cite{lu2023instag,li2023label}. Instruction Fine-tuning, a variant of SFT, focuses on teaching models to follow human instructions \cite{wang-etal-2023-self-instruct, xu2024wizardlm}, where LLMs are fine-tuned on large sets of instruction-response demonstrations either annotated by humans or synthesised by proprietary models. After instruction tuning, LLMs exhibit improved generalisation capabilities, even for previously unseen tasks \cite{sanh2022multitask,chung2024scaling}.

\subsubsection{Reinforcement Learning from Human Feedback (RLHF)}

RLHF optimises a model by incorporating human feedback during training \cite{Ouyang2022, OpenAI2022}. This technique improves the alignment of LLM outputs with human preferences and ethical standards \cite{rafailov2024direct,meng2024simpo}, particularly in complex environments where manually designing reward functions is challenging or insufficient. 

\subsection{External Interaction and Collaboration}

The techniques in this category focus on enhancing the interaction between LLMs and external elements such as databases, tools, environments, and user-provided information, thereby augmenting their capabilities.

\subsubsection{Prompt Engineering}

Prompt engineering involves crafting specific input prompts to elicit the desired responses from a LLM without altering its internal parameters \cite{liu2023pre,yao2024tree,besta2024graph}. Techniques like In-Context Learning (ICL) \cite{brown2020language,akyurek2023what} and Chain-of-Thought (CoT) \cite{wei2022chain} have been developed to improve the performance of LLMs on complex reasoning tasks. However, prompt engineering is limited by its reliance on prompt quality, especially in complex or domain-specific tasks.

\subsubsection{Retrieval-Augmented Generation (RAG)}

RAG is a technique that combines LLMs with external retrieval mechanisms, enabling the models to generate more accurate responses \cite{fan2024survey,ram2023context}. RAG-based methods follow the information indexing and retrieval, information augmentation, and answer generation paradigm \cite{guo-etal-2023-prompt, wang-etal-2023-self-knowledge}. This is especially useful when domain-specific knowledge is required, but the employed model itself may not have sufficient knowledge within its parameters or expertise on a certain topic. The integration of retrieval systems with LLMs allows for a dynamic access to up-to-date information, ensuring that the generated content reflects the most current and relevant data available \cite{zhang2023enhancing}.

\subsubsection{Agent-Based Methods}

Agent-based methods involve utilising LLMs as intelligent agents that interact with their environment to achieve specific goals. These agents typically consist of three key components: memory, planning, and execution \cite{wang2024survey, zhao2023survey, park2023}. A notable extension is represented by multi-agent systems, where multiple agents collaborate to solve tasks beyond the capacity of a single agent (Park et al., 2023; Li, G. et al., 2023)\cite{park2023,li2023a}. LLM-powered agents have been applied in various tasks, such as environment simulation \cite{park2023}, experiment automation \cite{odonoghue-etal-2023-bioplanner,yoshikawa2023large}, and education simulation \cite{zhang2024simulating}.

\subsubsection{Tool-Use Integration}

This technique integrates LLMs with external tools, such as specialised software or APIs, to enhance their ability to perform domain-specific tasks \cite{schick2023toolformer, patil2023gorilla,qin2024toolllm}. This paradigm allows LLMs to handle more complex, multi-step tasks such as data retrieval \cite{koldunov2024local,vaghefi2023chatclimate}, code execution \cite{Ma2024discovery,cai2024large}, and scientific simulations \cite{bran2023chemcrow,huang2024crispr}.

Table \ref{tab2} features the characteristics, advantages, limitations, and application scenarios of the techniques discussed, providing a clear comparison to aid in selecting the most suitable approach for specific tasks.

\begin{table*}[h]
\centering
\caption{Overview of the Techniques for Applying LLMs Across Various Disciplines}
\label{tab2}
\scalebox{0.78}{
\begin{tabular}{@{}cccccc@{}}
\toprule
\textbf{Category}  & \textbf{Technique}  & \textbf{Description}  & \textbf{Advantages} & \textbf{Limitations}  & \textbf{\begin{tabular}[c]{@{}c@{}}Application \\ Scenarios\end{tabular}} \\ \midrule
\multirow{3}{*}{\textit{\textbf{\begin{tabular}[c]{@{}c@{}}Internal \\ Knowledge\\ Optimisation\end{tabular}}}}            
& \begin{tabular}[c]{@{}c@{}}Continued \\ Pre-training\end{tabular}                    & \begin{tabular}[c]{@{}c@{}}Additional  training \\ on domain-specific \\ data to deepen \\ LLM expertise\end{tabular}   & \begin{tabular}[c]{@{}c@{}}Enhances  \\ domain-specific \\ knowledge\end{tabular}  & \begin{tabular}[c]{@{}c@{}}Requires massive  domain \\ data, involves a high \\ computational cost, \\ catastrophic forgetting\end{tabular}            & \begin{tabular}[c]{@{}c@{}}Scientific  research, \\ code generation, \\ mathematical reasoning\end{tabular}                  \\ 
\cmidrule(l){2-6} & \begin{tabular}[c]{@{}c@{}}Supervised \\ Fine-tuning\end{tabular}  
& \begin{tabular}[c]{@{}c@{}}Fine-tuning LLMs  \\ on labeled datasets \\ to specialise in \\ different tasks\end{tabular} & \begin{tabular}[c]{@{}c@{}}Fast adaptation  \\ to downstream \\ tasks, high \\ training efficiency\end{tabular} & \begin{tabular}[c]{@{}c@{}}Relies on  high-quality \\ labeled data, risk \\ overfitting and weakened \\ general capabilities\end{tabular}              & \begin{tabular}[c]{@{}c@{}}Task-oriented \\ applications \\ (e.g. sentiment \\ analysis, question \\ answering)\end{tabular} \\ 
\cmidrule(l){2-6} & \begin{tabular}[c]{@{}c@{}}Reinforcement \\ Learning \\ from Human \\ Feedback\end{tabular} & \begin{tabular}[c]{@{}c@{}}Collecting human  \\ feedback to refine \\ LLM behavior\end{tabular}  & \begin{tabular}[c]{@{}c@{}}Aligns model  \\ output with \\ human preferences\end{tabular}                       & \begin{tabular}[c]{@{}c@{}}Requires complex  \\ feedback mechanisms, \\ unstable training, high \\ resource consumption\end{tabular}                   & \begin{tabular}[c]{@{}c@{}}Ethics-focused \\ tasks, real-world \\ scenario \\ simulations\end{tabular}                       \\ \midrule
\multirow{4}{*}{\textit{\textbf{\begin{tabular}[c]{@{}c@{}}External \\ Interaction \\ and \\ Collaboration\end{tabular}}}} & 
\begin{tabular}[c]{@{}c@{}}Prompt \\ Engineering\end{tabular}  & \begin{tabular}[c]{@{}c@{}}Crafting input  \\ prompts to guide \\ model responses\end{tabular}                          & \begin{tabular}[c]{@{}c@{}}No training  \\ costs, high \\ flexibility\end{tabular}                              & \begin{tabular}[c]{@{}c@{}}Performance  dependent \\ on prompt design skills \\ and base model capability\end{tabular}                                 & \begin{tabular}[c]{@{}c@{}}Text generation, \\ translation, \\ question answering\end{tabular}   \\
\cmidrule(l){2-6} & \begin{tabular}[c]{@{}c@{}}Retrieval-\\ Augmented \\ Generation\end{tabular}                & \begin{tabular}[c]{@{}c@{}}Integrating  \\ external \\ knowledge \\ retrieval with \\ LLM generation\end{tabular}       & \begin{tabular}[c]{@{}c@{}}More accurate,  \\ up-to-date \\ responses\end{tabular}                              & \begin{tabular}[c]{@{}c@{}}Performance  depends \\ on retrieval quality, \\ increased latency, \\ requires knowledge \\ base  maintenance\end{tabular} & \begin{tabular}[c]{@{}c@{}}Medicine, law,  \\ finance, any \\ knowledge-\\ intensive field\end{tabular}                      \\ 
\cmidrule(l){2-6} & \begin{tabular}[c]{@{}c@{}}Agent-Based \\ Methods\end{tabular}       & \begin{tabular}[c]{@{}c@{}}LLMs interact  \\ with an \\ environment to \\ solve tasks\end{tabular}                      & \begin{tabular}[c]{@{}c@{}}Automates  \\ complex, \\ multi-step tasks\end{tabular}                              & \begin{tabular}[c]{@{}c@{}}Complex system design, \\ error propagation risks, \\ high resource demands\end{tabular}                                    & \begin{tabular}[c]{@{}c@{}}Task automation, \\ multi-agent \\ simulations\end{tabular} \\
\cmidrule(l){2-6} & \begin{tabular}[c]{@{}c@{}}Tool-Use \\ Integration\end{tabular}                             & \begin{tabular}[c]{@{}c@{}}Using external  \\ tools to extend \\ LLM capabilities\end{tabular}                          & \begin{tabular}[c]{@{}c@{}}Enables complex,  \\ multi-step \\ problem solving\end{tabular}                      & \begin{tabular}[c]{@{}c@{}}Requires tool  interface \\ development, potential \\ API failure risks\end{tabular}                                        & \begin{tabular}[c]{@{}c@{}}Scientific \\ simulation, \\ code generation, \\ Mathematical \\ computations\end{tabular}        \\ \bottomrule
\end{tabular}
}
\end{table*}

\section{LLM Applications in Different Disciplines} \label{sec4}

Automation tools and information systems have garnered extensive attention and application in the realms of data analysis and collaborative decision-making \cite{filip2021automation, filip2022collaborative}. LLMs have significantly impacted diverse research fields, driving advances in these fields and enabling novel interdisciplinary approaches. This section provides a comprehensive analysis of LLM applications in multiple disciplines, highlighting their contributions and vast potential in advancing discipline-specific research.

\subsection{Mathematics}

Mathematics, a foundational discipline, deals with complex, abstract problems that often require intricate reasoning and precise logic. Mathematical tasks ranging from algebra and calculus to combinatorics and theorem proving feature challenges that require advanced reasoning, logical consistency, and multi-step planning. LLMs have demonstrated their potential to aid in several areas of mathematical research.

\subsubsection{Mathematical Problem-Solving}

Mathematical problem-solving involves identifying relevant formulas, applying logical reasoning, and grounding solutions in mathematical principles. Recent studies have underscored the efficacy of LLMs in tackling mathematical problems \cite{shao2024deepseekmath,gou2024tora}. Fine-tuning LLMs on mathematical datasets has emerged as an effective strategy for enhancing their mathematical reasoning capabilities, yielding significant improvements on mathematical tasks \cite{lewkowycz2022solving, zhang2024sciglm}. Recent advancements, such as the integration of CoT and the utilisation of specialised tools, have improved their performance in intricate tasks \cite{wang2023selfconsistency,he2023solving}. Notably, \citet{jaech2024openai} reported that their latest model achieved a rank among the top 500 students in the U.S. with regard to the qualification for the USA Math Olympiad.

\subsubsection{Mathematical Theorem Proving}

By leveraging the extensive knowledge embedded in them, the LLMs have demonstrated their ability to generate proofs for mathematical theorems \cite{jiang2022thor, lample2022hypertree, yang2023leandojo}. Literature has explored decomposing proofs into simpler lemmas and formalising these structures \cite{wang2023lego}. \citet{yang2023leandojo} introduced an open-source platform, namely LeanDojo for theorem proving. However, a significant challenge in applying LLMs to mathematical theorem proving is the issue of hallucination.

\subsection{Physics}

Physics research, which demands not only logical reasoning but also the application of specific physical laws and experimental data \cite{bakhtin2019phyre, pang2024physics, jaiswal2024improving}, features unique challenges. LLMs have shown substantial potential in various physics-related tasks.

\subsubsection{Physical Hypothesis Generation}

LLMs can assist in generating new scientific hypotheses by synthesising large amounts of data and existing knowledge \cite{ciucua2023harnessing}. For instance, LLMs have been used to automatically extract governing equations from data and refine equations iteratively through reasoning \cite{meng2024equa}. \citet{Ma2024discovery} proposed a Scientific Generative Agent (SGA) to advance physical scientific discovery. In quantum mechanics, fine-tuned LLMs can generate hypotheses about quantum entanglement or propose experiments to test current models, opening new avenues for scientific discovery.

\subsubsection{Simulation Data Analysis}

In fields like astrophysics and particle physics, where experimental data is crucial for understanding complex phenomena \cite{wang2024can,sharma2025computational}, LLMs have been employed to assist in the analysis of large-scale simulation data. Agent-based methods, wherein LLMs function as autonomous agents interacting with simulated environments, have also proven particularly effective in these domains \cite{gao2024large}.

\subsubsection{Experimental Assistance}

LLMs have also been employed in experimental design, assisting physicists in developing novel experiments based on existing data and theories. \citet{mehta2023towards} proposed utilising LLMs with RAG for tokamak fusion reactors. By leveraging extensive text logs from past experiments, LLMs facilitate more efficient and data-driven experimental operations.

\subsection{Chemistry}

The powerful sequence processing capabilities of LLMs allow them to handle complex chemical terminologies and diverse sources of information, thereby advancing automation and intelligence in chemistry. LLMs have been applied across a wide range of aspects of chemistry, including molecular design, molecular property prediction, chemical reaction prediction, and chemical literature analysis \cite{fang2023mol,tang2024prioritizing}.

\subsubsection{Molecular Representations}

Researchers are actively exploring methods to leverage LLMs for constructing higher-quality molecular representations in both 2D and 3D formats through task-specific fine-tuning. \citet{fang2024molinstructions} introduced the Mol-Instruction dataset, demonstrating the potential of LLMs in molecular modeling. \citet{cao-etal-2025-instructmol} aligned 2D molecular structures with natural language via an instruction-tuning approach, while \citet{li2024towards} explored the advantages of 3D molecular representations in multimodal LLMs. PRESTO \cite{cao-etal-2024-presto} enhanced LLMs’ comprehension of molecular-related knowledge through extensive domain-specific pre-training.

\subsubsection{Molecular Design}

In molecular design, traditional methods rely heavily on expert knowledge and computationally intensive simulations. LLMs have introduced new possibilities in this field. \citet{fang2024domainagnostic} introduced MolGen, a pre-trained molecular language model specifically tailored for molecule generation. Similarly, \citet{frey2023nature} developed ChemGPT, a generative pre-trained Transformer model with over one billion parameters, for small molecule generation. These advancements have significantly streamlined the drug discovery process and the design of new materials.

\subsubsection{LLM Agents for Chemical Research}

The development of LLMs has led to advanced language agents that assist in chemical research \cite{ramos2025review}. ChemCrow \cite{m2024augmenting} integrates LLMs with chemical tools to perform a wide range of chemistry-related tasks, outperforming GPT-4 in accuracy. Coscientist \cite{boiko2023autonomous} combines semi-autonomous robots to plan and execute chemical reactions with minimal human intervention. Chemist-X \cite{2023arXiv231110776C} focuses on designing reactions for specific molecules, while ProtAgent \cite{ghafarollahi2024protagents} automates and optimises protein design. CACTUS \cite{mcnaughton2024cactus} automates the application of cheminformatics tools with human oversight in molecular discovery. These advancements highlight the transformative potential of LLM-powered agents in chemical research, with regard to streamlining processes, improving the efficiency of research, and accelerating scientific discovery.

\subsection{Biology}

LLMs have also made substantial contributions to biology. From protein structure prediction to genomic analysis, LLMs are increasingly being utilised to accelerate the understanding of biological systems, aid in drug discovery, and enhance the design of new biological molecules. This subsection examines how LLMs are revolutionising several areas of biology, particularly biological molecule analysis, protein analysis, and genomic research.

\subsubsection{Biological Molecular Analysis}

LLMs excel at analysing large-scale molecular data, helping to uncover hidden relationships and patterns, thereby providing valuable insights into molecular structures. For molecular representation, MoLFormer \cite{ross2022large} expands its pre-training dataset to 1.1 billion molecules, outperforming traditional self-supervised GNN methods. Beyond molecular data, integrating multimodal information such as targeting ligands, diseases, and other biochemical entities enhances the LLM models’ ability to capture comprehensive molecular knowledge. MolFM \cite{luo2023molfm} integrates molecular structures, biomedical texts, and knowledge graphs into a multimodal encoder, capturing both local and global molecular knowledge. BioMedGPT \cite{luo2023biomedgpt} further extends this by incorporating protein and biomedical data. These advancements highlight the promise of LLMs in providing a more comprehensive approach to biomolecular analysis by integrating diverse biological data.

\subsubsection{Protein Analysis}

LLMs have been extensively applied in protein-centric applications, including protein folding prediction \cite{jumper2021highly}, protein-protein interaction analysis \cite{jin2024prollm}, and function prediction \cite{zhang2023protein}. Protein-specific LLMs pre-trained on large-scale protein sequences, such as ProtLLM \cite{zhuo2024protllm}, ProLLM \cite{jin2024prollm}, and ProLLaMA \cite{lv2024prollama}, have emerged. These models have achieved an excellent performance in tasks such as protein sequence generation, protein property prediction, and other protein-related analyses.

\subsubsection{Genomic Sequence Analysis}

In genomics, LLMs are used to understand mutation effects and predict genomic features directly from DNA sequences. GenSLMs \cite{zvyagin2023genslms}, pre-trained on over 110 million gene sequences, has been applied to tasks like identifying genetic variants and modeling the SARS-CoV-2 genome. EpiBrainLLM \cite{liu2024genoagent} leverages genomic LLMs to improve the causal analysis from genotypes to brain measures and AD-related clinical phenotypes. These models enhance one’s ability to interpret genomic data, advancing personalised medicine and disease research.

\subsection{Humanities \& Social Sciences}

The application of LLMs has opened up new research frontiers in both Humanities and Social Sciences, enabling scholars to tackle longstanding challenges with a greater efficiency. This section explores specific application examples for LLMs in these two domains.

\subsubsection{LLMs in Humanities}

The humanities involve a profound exploration and reflection on human culture, thought, and values, with deep historical roots and rich cultural connotations. In the modern society, the humanities play an irreplaceable role in cultivating humanistic literacy, aesthetic abilities, critical thinking, and proper values. LLMs have multiple unique applications in the humanities, which can be broadly classified as follows.

\paragraph{\textbf{Textual Analysis and Interpretation.}}

In literary studies, LLMs conduct a detailed analysis across genres, capturing linguistic elements like vocabulary, grammar, and rhetoric to identify stylistic features \cite{beguvs2023large,beuls2024construction}. They can also track thematic evolution for various works across different eras, such as how the theme of love has changed in literature. By analysing word choices and syntactic structure, LLMs help attribute authorship and determine the compositional period for texts, supporting literary criticism.

\paragraph{\textbf{Cultural Heritage.}} 

LLMs are transforming cultural heritage \cite{filip2015cultural} through their capabilities to analyse documents \cite{zhang2025understand,zhang-etal-2025-chaotic,liang2024document}, to process and generate texts, and support multilingual translation (Li et al., 2024a; Li et al., 2024b)\cite{li-etal-2024-improving-context,DBLP:conf/acl/LiYZLWZ24}. They contribute to preserving low-resource languages by constructing linguistic corpora \cite{otieno2024framework}. Additionally, LLMs fine-tuned with specific cultural and historical datasets assist professionals in analysing historical texts and artifacts \cite{otieno2024framework}. LLMs enhance public engagement and education in cultural heritage, as seen in guided systems tailored to the visitors’ experiences \cite{trichopoulos2023large}.

\paragraph{\textbf{Cross-Cultural Research.}}

LLMs play a vital role in cross-cultural studies by analysing materials from various cultural backgrounds, helping researchers understand the differences and similarities between them. However, since LLMs are mainly trained on English language data, they may exhibit cultural bias, leading to stereotypical representations that exacerbate social conflicts \cite{masoud2023cultural,naous2023having}. Some efforts are focused on addressing this bias \cite{kim2024exploring,li2024culturepark}. Additionally, LLMs aid historians by extracting key information from vast amounts of historical material to construct a comprehensive view of historical events \cite{zeng2024histolens,garcia2023if}.

\subsubsection{LLMs in Social Sciences}

In Social Sciences, LLMs are employed to model human behavior, simulate social behaviours, predict social trends, and analyse sentiments. These capabilities are transforming research in areas like sociology, economics, political science, and psychology.

\paragraph{\textbf{Modeling Human Behavior.}}

LLMs are increasingly used in psychological experiments, providing certain advantages over traditional methods such as cost-effectiveness, scalability, and fewer ethical concerns \cite{griffin2023susceptibility}. They simulate the responses of individuals to stimuli, helping psychologists test hypotheses without human participants \cite{binz2023using,10.1093/pnasnexus/pgae245}. In light of this, various studies have compared LLMs with human participants from a behavioural perspective and demonstrate that LLMs align with human judgments \cite{argyle2023out}. This indicates their potential to model human individuals.

\paragraph{\textbf{Social behavior simulations.}}

LLMs are powerful tools in Computational Social Science, enabling the simulation of diverse scenarios and the study of emergent phenomena in controlled environments \cite{10.1162/coli_a_00502,bail2024can, gao2024large,ye-etal-2025-sweetiechat}. For instance, the CompeteAI framework simulates competition between LLM agents in a virtual town \cite{zhao2023competeai}, and EconAgent enhances macroeconomic simulations by modeling a more realistic decision-making process \cite{li-etal-2024-econagent}. AgentHospital simulates illness treatment processes using LLM-powered agents \cite{li2024agent}, while AgentReview addresses privacy concerns in peer review analysis \cite{jin2024agentreview}. Additionally, these simulations have been applied in the education domain \cite{zhang2024simulating}.

\paragraph{\textbf{Political Analysis and Sentiment Research}}

LLMs have revolutionised political science through the automation of extensive textual analysis, addressing challenges related to scalability, multilingual data, and unstructured texts \cite{zong2021text, heseltine2024large, najafi2024turkishbertweet, li2024political}. LLMs have been applied in various tasks, including political behavior analysis \cite{rozado2024political}, public opinion assessment \cite{breum2024persuasive}, election forecasting \cite{gujral2024can}, and sentiment analysis \cite{zhang-etal-2024-sentiment}. By automating these processes, LLMs have not only accelerated research but also enhanced its accuracy, enabling more thorough and reliable analyses \cite{wang2024intelligent}.

\section{Challenges and Future Directions} \label{sec5}

The application of LLMs across various disciplines has proven transformative, but several challenges remain that hinder them from reaching their full potential. Simultaneously, these challenges feature opportunities for innovation and outline the development of new research directions.

\subsection{Critical Challenges}

\subsubsection{Quality of Discipline-related datasets}

The performance of LLMs is highly dependent on the quality and diversity of their training data. For many discipline-specific tasks, LLMs often require either CPT or SFT on large, high-quality datasets. The need for substantial, domain-specific data becomes particularly evident when addressing highly specialised fields like medicine, chemistry, or physics, where a deep understanding of complex concepts is essential. However, obtaining large-scale, high-quality datasets for such tasks is a significant challenge. Furthermore, the process of curating and cleaning such data to ensure its relevance and quality can be resource-intensive.

\subsubsection{Usage Barriers for Non-AI Specialists}

For experts outside the AI field, utilising LLMs can feature significant challenges due to a lack of technical expertise in machine learning. This barrier limits the broader adoption of LLMs across various disciplines, as these experts may find it difficult to effectively integrate these models into their research workflows. Moreover, customising LLMs for specific domains often requires a substantial technical input, which may not be readily accessible in all disciplines.

\subsubsection{Lack of Standardised Evaluation Benchmarks}

The absence of universally accepted evaluation metrics and datasets for specific disciplines makes it difficult to assess the performance of LLMs in a standardised way. Different fields often rely on bespoke evaluation frameworks, which can lead to inconsistent comparisons and hinder cross-disciplinary collaboration.

\subsubsection{High Computational Costs}

Large-scale LLMs require substantial computational resources for their training and deployment. The associated costs, both in terms of hardware and energy consumption, represent significant barriers for many research institutions and smaller organisations. This limitation restricts the accessibility of LLMs, particularly in resource-constrained environments.

\subsection{Future Directions}

The challenges presented above highlight the need for targeted improvements in the application of LLMs across various disciplines. To address these issues, the following future directions can help overcome the current limitations and pave the way for a more effective integration of LLMs into scientific research.

\subsubsection{Improving the Access to High-Quality Discipline-Specific Data}

Future research should concentrate on developing efficient methods for generating and curating high-quality datasets across various disciplines. Collaborative initiatives involving domain experts and AI researchers can facilitate the creation of more diverse, accurate, and accessible datasets. Furthermore, leveraging techniques such as few-shot learning and transfer learning can optimise the use of limited discipline-specific data, thereby reducing the reliance on massive datasets.

\subsubsection{Facilitating Interdisciplinary Collaboration}

To bridge the gap between AI experts and discipline specialists, future efforts should emphasise the development of user-friendly tools and platforms that enable non-AI experts to easily access and apply LLMs. Tailored training programs for researchers in specific fields can equip domain experts with the necessary skills to utilise LLMs effectively. Additionally, creating collaborative platforms that facilitate interaction between AI practitioners and specialists from other fields can promote the integration of LLMs into a broader range of research areas.

\subsubsection{Developing Standardised Evaluation Metrics and Datasets}

Future work should focus on establishing standardised evaluation metrics and benchmark datasets tailored to specific disciplines. These standardised tools would enable researchers to assess a model’s performance more effectively, make meaningful comparisons across various studies, and ensure the reliability of the obtained results. The collaboration between domain experts and AI researchers will be crucial in developing evaluation frameworks that are both relevant and rigorous for each field.

\subsubsection{Reducing Computational Costs}

As LLMs continue to grow in size, it is essential to enhance their computational efficiency. Research into techniques such as model pruning, distillation, and other optimisation methods can help reduce the computational resources required for training and inference. Moreover, adopting energy-efficient hardware and exploring cloud-based solutions can increase the accessibility of LLMs for a wider range of research institutions.

\section{Conclusion} \label{sec6}

This survey provides a comprehensive overview of the applications of LLMs across various academic disciplines, highlighting both their transformative potential and the associated challenges. By systematically categorising LLM techniques, a novel principled taxonomy which elucidates how these models enhance research across different disciplines was introduced. Further on, the practical applications of LLMs in mathematics, physics, chemistry, biology, and the humanities and social sciences are analysed, which demonstrates their ability to facilitate complex problem-solving, knowledge synthesis, and data-driven discoveries. While LLMs have shown remarkable capabilities in assisting discipline-specific research, they also feature significant challenges, including discipline-specific data limitations, usage barriers for non-AI specialists, the lack of standardised evaluation benchmarks, and high computational costs. Addressing these challenges requires concerted efforts to improve the access to high-quality domain-specific data, to facilitate interdisciplinary collaboration, develop standardised evaluation metrics and datasets, and reduce computational costs. This survey underscores the necessity of interdisciplinary collaboration to advance the use of LLMs in discipline-specific research. By fostering this kind of collaboration, LLM methodologies can be refined to better align with discipline-specific needs, ultimately leading to more informed, accurate, and innovative scientific contributions.

\section*{Acknowledgments}

This paper is partially supported by the Natural 
Science Foundation of China under Grant No. 
62336008. Since the authors met Academician 
Prof. F. G. Filip from the Romanian Academy 
in November 2024, they have maintained 
close contact and cooperation. Prof. F. G. Filip 
has given us substantial encouragement and 
a lot of suggestions regarding the research in 
artificial intelligence and LLMs for information 
processing and other applications. His wisdom 
inspired the authors to write this article in order 
to comprehensively analyse and summarise the 
current R\&D status for LLMs in interdisciplinary 
fields. The authors sincerely appreciate his 
encouragement and support!

\bibliography{custom}

\end{document}

%% file: 00_definitions.tex
\definecolor{green}{rgb}{0.1,0.1,0.1}
\definecolor{chocolate}{HTML}{D2691E}
\definecolor{maroon}{HTML}{A00000}
\definecolor{indigo}{HTML}{4B0082}
\definecolor{green}{HTML}{008000}
\definecolor{cadmiumgreen}{rgb}{0.0, 0.42, 0.24}

\definecolor{airforceblue}{rgb}{0.36, 0.54, 0.66}
\definecolor{Gray}{gray}{0.9}

\usepackage[edges]{forest}
\definecolor{lightcoral}{rgb}{0.94, 0.5, 0.5}
\definecolor{lightgreen}{rgb}{0.56, 0.93, 0.56}

\definecolor{brightlavender}{rgb}{0.75, 0.58, 0.89}
\definecolor{capri}{rgb}{0.0, 0.75, 1.0}
\definecolor{carminepink}{rgb}{0.92, 0.3, 0.26}
\definecolor{celadon}{rgb}{0.67, 0.88, 0.69}
\definecolor{darkpastelgreen}{rgb}{0.01, 0.75, 0.24}

\definecolor{pastelblue}{rgb}{0.68, 0.78, 0.81}
\definecolor{mintgreen}{rgb}{0.6, 0.98, 0.6}
\definecolor{lavender}{rgb}{0.71, 0.49, 0.86}
\definecolor{peach}{rgb}{1.0, 0.9, 0.71}
\definecolor{coral}{rgb}{1.0, 0.5, 0.31}
\definecolor{mauve}{rgb}{0.88, 0.69, 1.0}
\definecolor{lemonyellow}{rgb}{1.0, 0.96, 0.4}

\definecolor{hidden-draw}{RGB}{205, 44, 36}
\definecolor{hidden-blue}{RGB}{194,232,247}
\definecolor{hidden-orange}{RGB}{243,202,120}
\definecolor{hidden-yellow}{RGB}{242,244,193}
\definecolor{tree-level-1}{RGB}{245,20,85}
\definecolor{tree-level-2}{RGB}{246,86,118}
\definecolor{tree-level-3}{RGB}{248,177,193}
\definecolor{tree-leaf}{RGB}{176,230,198}

\definecolor{Self}{RGB}{255,0,128}
\definecolor{Ensemble}{RGB}{0,127,255}
\definecolor{Iterative}{RGB}{153,51,255}

\definecolor{exemplar1}{RGB}{136,98,148}
\definecolor{exemplar2}{RGB}{148,210,242}
\definecolor{knowledge1}{RGB}{249,219,152}
\definecolor{knowledge2}{RGB}{255,245,220}

\definecolor{lighttealblue}{RGB}{41, 157, 143}
\definecolor{lightplum}{RGB}{233, 196, 106}
\definecolor{harvestgold}{RGB}{216, 118, 89}


%% file: Figures/category_tree.tex
\tikzstyle{my-box}=[
    rectangle,
    rounded corners,
    text opacity=1,
    minimum height=1.5em,
    minimum width=5em,
    inner sep=2pt,
    align=center,
    fill opacity=.5,
]
\tikzstyle{hypothesis_leaf}=[my-box, minimum height=1.5em,
    fill=cyan!20, text=black, align=left,font=\small,
    inner xsep=2pt,
    inner ysep=4pt,
]
\tikzstyle{cause_leaf}=[my-box, minimum height=1.5em,
    fill=lighttealblue!20, text=black, align=left,font=\small,
    inner xsep=2pt,
    inner ysep=4pt,
]
\tikzstyle{detect_leaf}=[my-box, minimum height=1.5em,
    fill=lightplum!20, text=black, align=left,font=\small,
    inner xsep=2pt,
    inner ysep=4pt,
]
\tikzstyle{mitigate_leaf}=[my-box, minimum height=1.5em,
    fill=harvestgold!20, text=black, align=left,font=\small,
    inner xsep=2pt,
    inner ysep=4pt,
]

\begin{figure*}[!ht]
    \centering
    \resizebox{\textwidth}{!}{%
    \begin{forest}
        forked edges,
        for tree={
            grow=east,
            reversed=true,
            anchor=base west,
            parent anchor=east,
            child anchor=west,
            base=left,
            font=\normalsize,
            rectangle,
            rounded corners,
            align=left,
            minimum width=4em,
            edge+={darkgray, line width=1pt},
            s sep=3pt,
            inner xsep=2pt,
            inner ysep=3pt,
            ver/.style={rotate=90, child anchor=north, parent anchor=south, anchor=center},
        },
        where level=1{text width=7.5em,font=\normalsize}{},
        where level=2{text width=5.5em,font=\normalsize}{},
        where level=3{text width=6.5em,font=\small}{},
        where level=4{text width=7.5em,font=\small}{},
            [
                \textbf{LLM4Discipline}, color=carminepink!100, fill=carminepink!15, text=black
                [
                    \textit{Internal} \\ \textit{Knowledge} \\ \textit{Optimization}, color=cyan!100, fill=cyan!100, text=black
                    [
                        CPT, color=cyan!100, fill=cyan!60, text=black
                        [
                            Mathematics, color=cyan!100, fill=cyan!40, text=black
                            [
                                {\textsc{Llemma}~\citep{azerbayev2023llemma}, \textsc{Deepseekmath}~\citep{shao2024deepseekmath}}, hypothesis_leaf, text width=29.1em 
                            ] 
                        ]
                        [
                            Physics, color=cyan!100, fill=cyan!40, text=black
                            [
                                {\textsc{AstroMLab 2} \cite{pan2024astromlab}}, hypothesis_leaf, text width=29.1em   
                            ] 
                        ]
                    ]
                    [
                        SFT, color=cyan!100, fill=cyan!60, text=black
                        [
                            Mathematics, color=cyan!100, fill=cyan!40, text=black
                            [
                                {
                                \textsc{PaLM 2-L-Math} \cite{liu2023improving}, \textsc{Qwen-Math} \cite{yang2024qwen2_math}
                                }, hypothesis_leaf, text width=29.1em   
                            ] 
                        ]
                        [
                            Chemistry, color=cyan!100, fill=cyan!40, text=black
                            [
                                { \textsc{InstructMol} \cite{cao-etal-2025-instructmol},  \textsc{Mol-Instructions} \cite{fang2023mol}, \\
                                \textsc{ChemDFM} \cite{zhao2024chemdfm}
                                }, hypothesis_leaf, text width=29.1em   
                            ] 
                        ]
                        [
                            Physics, color=cyan!100, fill=cyan!40, text=black
                            [
                                { 
                                \textsc{AstroLLaMA} \cite{nguyen2023astrollama}, \\
                                \textsc{AstroLLaMA-Chat} \cite{perkowski2024astrollama}
                                }, hypothesis_leaf, text width=29.1em   
                            ] 
                        ]
                        [
                            Biology, color=cyan!100, fill=cyan!40, text=black
                            [
                                {  \textsc{PMC-LLaMA} \cite{wu2024pmc}, \textsc{ProLLaMA} \cite{lv2024prollama}
                                }, hypothesis_leaf, text width=29.1em   
                            ] 
                        ]
                    ]
                    [
                        RLHF, color=cyan!100, fill=cyan!60, text=black
                        [
                            Mathematics, color=cyan!100, fill=cyan!40, text=black
                            [
                                {
                                    \textsc{Deepseek-prover} \cite{xin2024deepseek}
                                }, hypothesis_leaf, text width=29.1em   
                            ] 
                        ]
                        [
                            Chemistry, color=cyan!100, fill=cyan!40, text=black
                            [
                                {
                                    \textsc{Bindgpt} \cite{zholus2024bindgpt}
                                }, hypothesis_leaf, text width=29.1em   
                            ] 
                        ]
                        [
                            Physics, color=cyan!100, fill=cyan!40, text=black
                            [
                                {
                                    \textsc{RLHAIF} \cite{anand2024enhancing}
                                }, hypothesis_leaf, text width=29.1em   
                            ] 
                        ]
                    ]
                ]
               [
                    \textit{External} \\ \textit{Interaction} \\ \textit{and} \\ \textit{Collaboration}, color=harvestgold!100, fill=harvestgold!100, text=black
                    [
                        Prompt \\Engineering, color=harvestgold!100, fill=harvestgold!60, text=black
                        [
                            Mathematics, color=harvestgold!100, fill=harvestgold!20, text=black
                            [
                                {
                                \textsc{codex-math} \cite{drori2022neural},   \textsc{SymbCoT} \cite{xu2024faithful}},
                                mitigate_leaf, text width=29.1em
                            ]
                        ]
                        [
                            Physics, color=harvestgold!100, fill=harvestgold!20, text=black
                            [
                                {
                                \textsc{Astro-GPT} \cite{ciucua2023harnessing}, \textsc{Mm-phyqa} \cite{anand2024mm}
                                },
                                mitigate_leaf, text width=29.1em
                            ]
                        ]
                        [
                            Biology, color=harvestgold!100, fill=harvestgold!20, text=black
                            [
                                {
                                scReader \cite{li2024screader}
                                },
                                mitigate_leaf, text width=29.1em
                            ]
                        ]
                        [
                            Chemistry, color=harvestgold!100, fill=harvestgold!20, text=black
                            [
                                {
                                \textsc{domain-knowledge embedded prompt engineering} \cite{liu2025integrating}
                                },
                                mitigate_leaf, text width=29.1em
                            ]
                        ]
                        [
                            Humanities \& \\Social Sciences, color=harvestgold!100, fill=harvestgold!20, text=black
                            [
                                {
                                \textsc{HistoLens} \cite{zeng2024histolens}, \textsc{PR-CoT} \cite{wang2023enhance}, \\ \textsc{HiSS} \cite{zhang2023towards}
                                },
                                mitigate_leaf, text width=29.1em
                            ]
                        ]
                    ]
                    [
                        RAG, color=harvestgold!100, fill=harvestgold!60, text=black
                        [
                            Mathematics, color=harvestgold!100, fill=harvestgold!20, text=black
                            [
                                {
                                \textsc{Leandojo} \cite{yang2023leandojo}
                                },
                                mitigate_leaf, text width=29.1em
                            ]
                        ]
                        [
                            Physics, color=harvestgold!100, fill=harvestgold!20, text=black
                            [
                                {
                                \textsc{SciPhy-RAG} \cite{anand2023sciphyrag}
                                },
                                mitigate_leaf, text width=29.1em
                            ]
                        ]
                        [
                            Biology, color=harvestgold!100, fill=harvestgold!20, text=black
                            [
                                {
                                \textsc{BioRAG} \cite{wang2024biorag}, \textsc{Self-BioRAG} \cite{jeong2024improving}
                                },
                                mitigate_leaf, text width=29.1em
                            ]
                        ]
                        [
                            Chemistry, color=harvestgold!100, fill=harvestgold!20, text=black
                            [
                                {
                                \textsc{ChatGPT Chemistry Assistant} \cite{zheng2023chatgpt}, \\
                                \textsc{DRAK-K} \cite{liu2024drak}
                                },
                                mitigate_leaf, text width=29.1em
                            ]
                        ]
                        [
                            Humanities \& \\Social Sciences, color=harvestgold!100, fill=harvestgold!20, text=black
                            [
                                {
                                \textsc{RAGAR} \cite{khaliq2024ragar}
                                },
                                mitigate_leaf, text width=29.1em
                            ]
                        ]
                    ]
                    [
                        Agent-based \\Methods, color=harvestgold!100, fill=harvestgold!60, text=black
                        [
                           Chemistry, color=harvestgold!100, fill=harvestgold!20, text=black
                            [
                                {
                                \textsc{Coscientist} \cite{boiko2023autonomous}, \textsc{CLAIRify} \cite{yoshikawa2023large}, \\
                                \textsc{ORGANA} \cite{darvish2025organa}
                                },
                                mitigate_leaf, text width=29.1em
                            ]
                        ]
                        [
                           Physics, color=harvestgold!100, fill=harvestgold!20, text=black
                            [
                                {
                                \textsc{StarWhisper Telescope} \cite{wang2024starwhisper}
                                },
                                mitigate_leaf, text width=29.1em
                            ]
                        ]
                        [
                           Biology, color=harvestgold!100, fill=harvestgold!20, text=black
                            [
                                {
                                \textsc{CellAgent} \cite{xiao2024cellagent}, \textsc{biomedical AI agents} \cite{gao2024empowering}, \\
                                \textsc{ProtAgents} \cite{ghafarollahi2024protagents}
                                },
                                mitigate_leaf, text width=29.1em
                            ]
                        ]
                        [
                           Humanities \& \\Social Sciences, color=harvestgold!100, fill=harvestgold!20, text=black
                            [
                                {
                                \textsc{CompeteAI} \cite{zhao2023competeai}, \textsc{Agent Hospital} \cite{li2024agent}, \\
                                \textsc{AgentReview} \cite{jin2024agentreview}
                                },
                                mitigate_leaf, text width=29.1em
                            ]
                        ]
                    ]
                    [
                        Tool-use \\Integration, color=harvestgold!100, fill=harvestgold!60, text=black
                        [
                           Mathematics, color=harvestgold!100, fill=harvestgold!20, text=black
                            [
                                {
                                \textsc{Toolformer} \cite{schick2023toolformer}, \textsc{ToRA} \cite{gou2024tora},  \\
                                \textsc{Code-scratchpad} \cite{upadhyay2023improving}, \textsc{LeanReasoner} \cite{jiang-etal-2024-leanreasoner}
                                },
                                mitigate_leaf, text width=29.1em
                            ]
                        ]
                        [
                           Chemistry, color=harvestgold!100, fill=harvestgold!20, text=black
                            [
                                {
                                \textsc{ChemCrow} \cite{bran2023chemcrow}
                                },
                                mitigate_leaf, text width=29.1em
                            ]
                        ]
                        [
                           Biology, color=harvestgold!100, fill=harvestgold!20, text=black
                            [
                                {
                                \textsc{GeneGPT} \cite{btae075_jin}, \textsc{CRISPR-GPT} \cite{huang2024crispr} 
                                },
                                mitigate_leaf, text width=29.1em
                            ]
                        ]
                    ]
                ]
            ]
    \end{forest}
    }
    \caption{Taxonomy of LLM techniques for different disciplines.}
    \label{fig:categorization_of_survey}
\end{figure*}
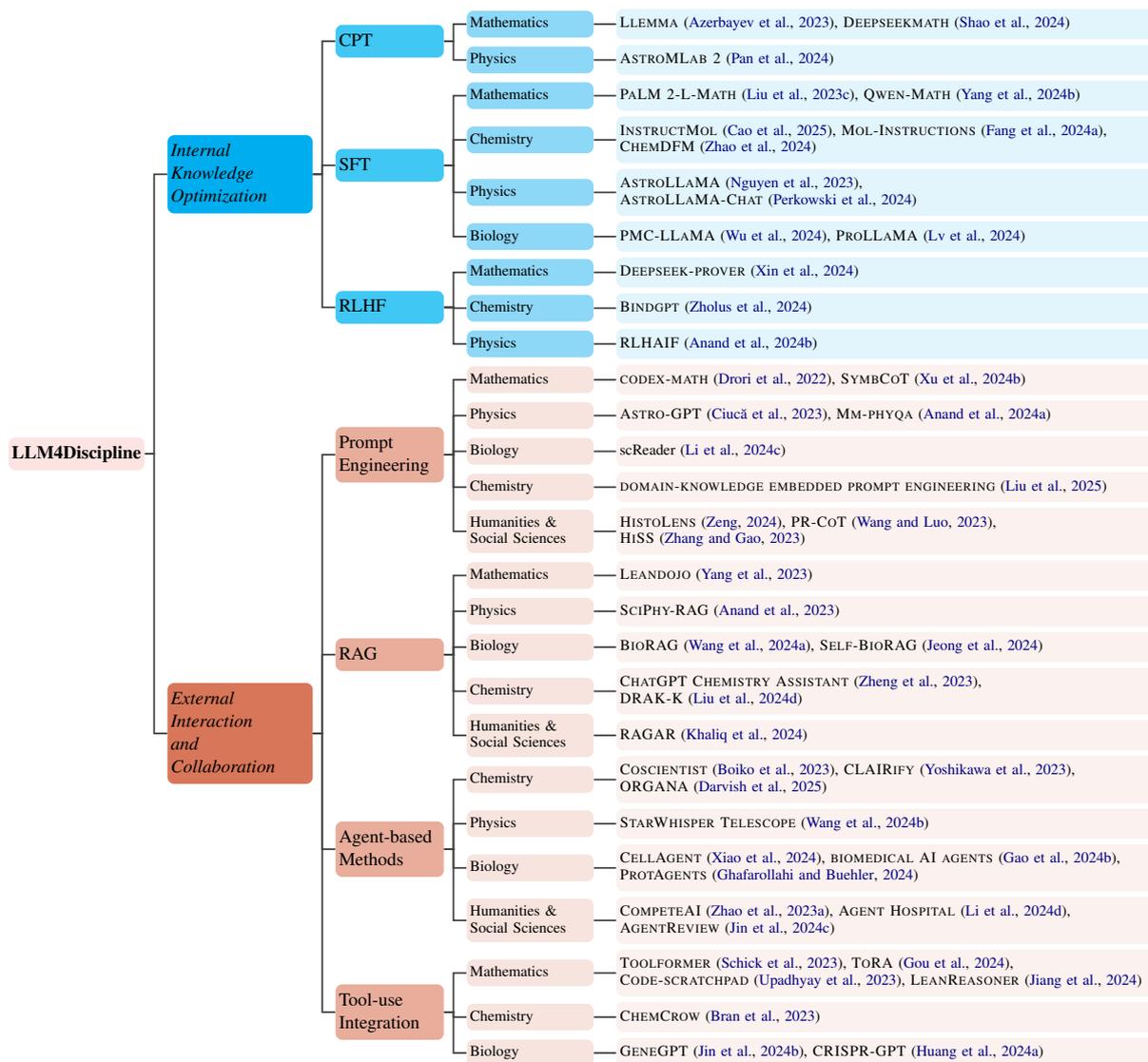

%% file: main.bbl
\begin{thebibliography}{173}
\providecommand{\natexlab}[1]{#1}

\bibitem[{Abdurahman et~al.(2024)Abdurahman, Atari, Karimi-Malekabadi, Xue, Trager, Park, Golazizian, Omrani, and Dehghani}]{10.1093/pnasnexus/pgae245}
Suhaib Abdurahman, Mohammad Atari, Farzan Karimi-Malekabadi, Mona~J Xue, Jackson Trager, Peter~S Park, Preni Golazizian, Ali Omrani, and Morteza Dehghani. 2024.
\newblock \href {https://doi.org/10.1093/pnasnexus/pgae245} {Perils and opportunities in using large language models in psychological research}.
\newblock \emph{PNAS Nexus}, 3(7):pgae245.

\bibitem[{Aky{\"u}rek et~al.(2023)Aky{\"u}rek, Schuurmans, Andreas, Ma, and Zhou}]{akyurek2023what}
Ekin Aky{\"u}rek, Dale Schuurmans, Jacob Andreas, Tengyu Ma, and Denny Zhou. 2023.
\newblock \href {https://openreview.net/forum?id=0g0X4H8yN4I} {What learning algorithm is in-context learning? investigations with linear models}.
\newblock In \emph{The Eleventh International Conference on Learning Representations}.

\bibitem[{Anand et~al.(2023)Anand, Goel, Hira, Buldeo, Kumar, Verma, Gupta, and Shah}]{anand2023sciphyrag}
Avinash Anand, Arnav Goel, Medha Hira, Snehal Buldeo, Jatin Kumar, Astha Verma, Rushali Gupta, and Rajiv~Ratn Shah. 2023.
\newblock Sciphyrag-retrieval augmentation to improve llms on physics q \&a.
\newblock In \emph{International Conference on Big Data Analytics}, pages 50--63. Springer.

\bibitem[{Anand et~al.(2024{\natexlab{a}})Anand, Kapuriya, Singh, Saraf, Lal, Verma, Gupta, and Shah}]{anand2024mm}
Avinash Anand, Janak Kapuriya, Apoorv Singh, Jay Saraf, Naman Lal, Astha Verma, Rushali Gupta, and Rajiv Shah. 2024{\natexlab{a}}.
\newblock Mm-phyqa: Multimodal physics question-answering with multi-image cot prompting.
\newblock In \emph{Pacific-Asia Conference on Knowledge Discovery and Data Mining}, pages 53--64. Springer.

\bibitem[{Anand et~al.(2024{\natexlab{b}})Anand, Prasad, Kirtani, Nair, Gupta, Garg, Gautam, Buldeo, and Shah}]{anand2024enhancing}
Avinash Anand, Kritarth Prasad, Chhavi Kirtani, Ashwin~R Nair, Mohit Gupta, Saloni Garg, Anurag Gautam, Snehal Buldeo, and Rajiv~Ratn Shah. 2024{\natexlab{b}}.
\newblock Enhancing llms for physics problem-solving using reinforcement learning with human-ai feedback.
\newblock \emph{arXiv preprint arXiv:2412.06827}.

\bibitem[{Argyle et~al.(2023)Argyle, Busby, Fulda, Gubler, Rytting, and Wingate}]{argyle2023out}
Lisa~P Argyle, Ethan~C Busby, Nancy Fulda, Joshua~R Gubler, Christopher Rytting, and David Wingate. 2023.
\newblock Out of one, many: Using language models to simulate human samples.
\newblock \emph{Political Analysis}, 31(3):337--351.

\bibitem[{Azerbayev et~al.(2023)Azerbayev, Schoelkopf, Paster, Santos, McAleer, Jiang, Deng, Biderman, and Welleck}]{azerbayev2023llemma}
Zhangir Azerbayev, Hailey Schoelkopf, Keiran Paster, Marco~Dos Santos, Stephen McAleer, Albert~Q Jiang, Jia Deng, Stella Biderman, and Sean Welleck. 2023.
\newblock Llemma: An open language model for mathematics.
\newblock \emph{arXiv preprint arXiv:2310.10631}.

\bibitem[{Bai et~al.(2023)Bai, Bai, Chu, Cui, Dang, Deng, Fan, Ge, Han, Huang et~al.}]{bai2023qwen}
Jinze Bai, Shuai Bai, Yunfei Chu, Zeyu Cui, Kai Dang, Xiaodong Deng, Yang Fan, Wenbin Ge, Yu~Han, Fei Huang, et~al. 2023.
\newblock Qwen technical report.
\newblock \emph{arXiv preprint arXiv:2309.16609}.

\bibitem[{Bail(2024)}]{bail2024can}
Christopher~A Bail. 2024.
\newblock Can generative ai improve social science?
\newblock \emph{Proceedings of the National Academy of Sciences}, 121(21):e2314021121.

\bibitem[{Bakhtin et~al.(2019)Bakhtin, van~der Maaten, Johnson, Gustafson, and Girshick}]{bakhtin2019phyre}
Anton Bakhtin, Laurens van~der Maaten, Justin Johnson, Laura Gustafson, and Ross Girshick. 2019.
\newblock Phyre: A new benchmark for physical reasoning.
\newblock \emph{Advances in Neural Information Processing Systems}, 32.

\bibitem[{Begu{\v{s}} et~al.(2023)Begu{\v{s}}, D{\k{a}}bkowski, and Rhodes}]{beguvs2023large}
Ga{\v{s}}per Begu{\v{s}}, Maksymilian D{\k{a}}bkowski, and Ryan Rhodes. 2023.
\newblock Large linguistic models: Analyzing theoretical linguistic abilities of llms.
\newblock \emph{arXiv preprint arXiv:2305.00948}.

\bibitem[{Besta et~al.(2024)Besta, Blach, Kubicek, Gerstenberger, Podstawski, Gianinazzi, Gajda, Lehmann, Niewiadomski, Nyczyk et~al.}]{besta2024graph}
Maciej Besta, Nils Blach, Ales Kubicek, Robert Gerstenberger, Michal Podstawski, Lukas Gianinazzi, Joanna Gajda, Tomasz Lehmann, Hubert Niewiadomski, Piotr Nyczyk, et~al. 2024.
\newblock Graph of thoughts: Solving elaborate problems with large language models.
\newblock In \emph{Proceedings of the AAAI Conference on Artificial Intelligence}, volume~38, pages 17682--17690.

\bibitem[{Beuls and Van~Eecke(2024)}]{beuls2024construction}
Katrien Beuls and Paul Van~Eecke. 2024.
\newblock Construction grammar and artificial intelligence.
\newblock In \emph{The Cambridge Handbook of Construction Grammar}. Cambridge University Press.

\bibitem[{Bi et~al.(2024)Bi, Chen, Chen, Chen, Dai, Deng, Ding, Dong, Du, Fu et~al.}]{bi2024deepseek}
Xiao Bi, Deli Chen, Guanting Chen, Shanhuang Chen, Damai Dai, Chengqi Deng, Honghui Ding, Kai Dong, Qiushi Du, Zhe Fu, et~al. 2024.
\newblock Deepseek llm: Scaling open-source language models with longtermism.
\newblock \emph{arXiv preprint arXiv:2401.02954}.

\bibitem[{Binz and Schulz(2023)}]{binz2023using}
Marcel Binz and Eric Schulz. 2023.
\newblock Using cognitive psychology to understand gpt-3.
\newblock \emph{Proceedings of the National Academy of Sciences}, 120(6):e2218523120.

\bibitem[{Boiko et~al.(2023)Boiko, MacKnight, Kline, and Gomes}]{boiko2023autonomous}
Daniil~A Boiko, Robert MacKnight, Ben Kline, and Gabe Gomes. 2023.
\newblock Autonomous chemical research with large language models.
\newblock \emph{Nature}, 624(7992):570--578.

\bibitem[{Bran et~al.(2023)Bran, Cox, Schilter, Baldassari, White, and Schwaller}]{bran2023chemcrow}
Andres~M Bran, Sam Cox, Oliver Schilter, Carlo Baldassari, Andrew~D White, and Philippe Schwaller. 2023.
\newblock Chemcrow: Augmenting large-language models with chemistry tools.
\newblock \emph{arXiv preprint arXiv:2304.05376}.

\bibitem[{Breum et~al.(2024)Breum, Egdal, Mortensen, M{\o}ller, and Aiello}]{breum2024persuasive}
Simon~Martin Breum, Daniel~V{\ae}dele Egdal, Victor~Gram Mortensen, Anders~Giovanni M{\o}ller, and Luca~Maria Aiello. 2024.
\newblock The persuasive power of large language models.
\newblock In \emph{Proceedings of the International AAAI Conference on Web and Social Media}, volume~18, pages 152--163.

\bibitem[{Brown et~al.(2020)Brown, Mann, Ryder, Subbiah, Kaplan, Dhariwal, Neelakantan, Shyam, Sastry, Askell et~al.}]{brown2020language}
Tom Brown, Benjamin Mann, Nick Ryder, Melanie Subbiah, Jared~D Kaplan, Prafulla Dhariwal, Arvind Neelakantan, Pranav Shyam, Girish Sastry, Amanda Askell, et~al. 2020.
\newblock Language models are few-shot learners.
\newblock \emph{Advances in neural information processing systems}, 33:1877--1901.

\bibitem[{Cai et~al.(2024)Cai, Wang, Ma, Chen, and Zhou}]{cai2024large}
Tianle Cai, Xuezhi Wang, Tengyu Ma, Xinyun Chen, and Denny Zhou. 2024.
\newblock \href {https://openreview.net/forum?id=qV83K9d5WB} {Large language models as tool makers}.
\newblock In \emph{The Twelfth International Conference on Learning Representations}.

\bibitem[{Cao et~al.(2025)Cao, Liu, Lu, Yao, and Li}]{cao-etal-2025-instructmol}
He~Cao, Zijing Liu, Xingyu Lu, Yuan Yao, and Yu~Li. 2025.
\newblock \href {https://aclanthology.org/2025.coling-main.25/} {{I}nstruct{M}ol: Multi-modal integration for building a versatile and reliable molecular assistant in drug discovery}.
\newblock In \emph{Proceedings of the 31st International Conference on Computational Linguistics}, pages 354--379, Abu Dhabi, UAE. Association for Computational Linguistics.

\bibitem[{Cao et~al.(2024)Cao, Shao, Liu, Liu, Tang, Yao, and Li}]{cao-etal-2024-presto}
He~Cao, Yanjun Shao, Zhiyuan Liu, Zijing Liu, Xiangru Tang, Yuan Yao, and Yu~Li. 2024.
\newblock \href {https://doi.org/10.18653/v1/2024.findings-emnlp.597} {{PRESTO}: Progressive pretraining enhances synthetic chemistry outcomes}.
\newblock In \emph{Findings of the Association for Computational Linguistics: EMNLP 2024}, pages 10197--10224, Miami, Florida, USA. Association for Computational Linguistics.

\bibitem[{{Chen} et~al.(2023){Chen}, {Li}, {Wang}, {Du}, {Yu}, {Lu}, {Li}, {Qiu}, {Pan}, {Huang}, {Fang}, {Heng}, and {Chen}}]{2023arXiv231110776C}
Kexin {Chen}, Junyou {Li}, Kunyi {Wang}, Yuyang {Du}, Jiahui {Yu}, Jiamin {Lu}, Lanqing {Li}, Jiezhong {Qiu}, Jianzhang {Pan}, Yi~{Huang}, Qun {Fang}, Pheng~Ann {Heng}, and Guangyong {Chen}. 2023.
\newblock \href {https://doi.org/10.48550/arXiv.2311.10776} {{Chemist-X: Large Language Model-empowered Agent for Reaction Condition Recommendation in Chemical Synthesis}}.
\newblock \emph{arXiv e-prints}, arXiv:2311.10776.

\bibitem[{Chowdhery et~al.(2023)Chowdhery, Narang, Devlin, Bosma, Mishra, Roberts, Barham et~al.}]{Chowdhery2023}
Aakanksha Chowdhery, Sharan Narang, Jacob Devlin, Maarten Bosma, Gaurav Mishra, Adam Roberts, Paul Barham, et~al. 2023.
\newblock Palm: scaling language modeling with pathways.
\newblock \emph{J. Mach. Learn. Res.}, 24(1).

\bibitem[{Chung et~al.(2024)Chung, Hou, Longpre, Zoph, Tay, Fedus, Li, Wang, Dehghani, Brahma et~al.}]{chung2024scaling}
Hyung~Won Chung, Le~Hou, Shayne Longpre, Barret Zoph, Yi~Tay, William Fedus, Yunxuan Li, Xuezhi Wang, Mostafa Dehghani, Siddhartha Brahma, et~al. 2024.
\newblock Scaling instruction-finetuned language models.
\newblock \emph{Journal of Machine Learning Research}, 25(70):1--53.

\bibitem[{Ciuc{\u{a}} et~al.(2023)Ciuc{\u{a}}, Ting, Kruk, and Iyer}]{ciucua2023harnessing}
Ioana Ciuc{\u{a}}, Yuan-Sen Ting, Sandor Kruk, and Kartheik Iyer. 2023.
\newblock Harnessing the power of adversarial prompting and large language models for robust hypothesis generation in astronomy.
\newblock \emph{arXiv preprint arXiv:2306.11648}.

\bibitem[{Darvish et~al.(2025)Darvish, Skreta, Zhao, Yoshikawa, Som, Bogdanovic, Cao, Hao, Xu, Aspuru-Guzik et~al.}]{darvish2025organa}
Kourosh Darvish, Marta Skreta, Yuchi Zhao, Naruki Yoshikawa, Sagnik Som, Miroslav Bogdanovic, Yang Cao, Han Hao, Haoping Xu, Al{\'a}n Aspuru-Guzik, et~al. 2025.
\newblock Organa: a robotic assistant for automated chemistry experimentation and characterization.
\newblock \emph{Matter}, 8(2).

\bibitem[{Devlin et~al.(2019)Devlin, Chang, Lee, and Toutanova}]{devlin-etal-2019-bert}
Jacob Devlin, Ming-Wei Chang, Kenton Lee, and Kristina Toutanova. 2019.
\newblock \href {https://doi.org/10.18653/v1/N19-1423} {{BERT}: Pre-training of deep bidirectional transformers for language understanding}.
\newblock In \emph{Proceedings of the 2019 Conference of the North {A}merican Chapter of the Association for Computational Linguistics: Human Language Technologies, Volume 1 (Long and Short Papers)}, pages 4171--4186, Minneapolis, Minnesota. Association for Computational Linguistics.

\bibitem[{Drori et~al.(2022)Drori, Zhang, Shuttleworth, Tang, Lu, Ke, Liu, Chen, Tran, Cheng et~al.}]{drori2022neural}
Iddo Drori, Sarah Zhang, Reece Shuttleworth, Leonard Tang, Albert Lu, Elizabeth Ke, Kevin Liu, Linda Chen, Sunny Tran, Newman Cheng, et~al. 2022.
\newblock A neural network solves, explains, and generates university math problems by program synthesis and few-shot learning at human level.
\newblock \emph{Proceedings of the National Academy of Sciences}, 119(32):e2123433119.

\bibitem[{Du et~al.(2024)Du, Chen, Wang, Nie, and Zhang}]{meng2024equa}
Mengge Du, Yuntian Chen, Zhongzheng Wang, Longfeng Nie, and Dongxiao Zhang. 2024.
\newblock \href {https://doi.org/10.1063/5.0224297} {Large language models for automatic equation discovery of nonlinear dynamics}.
\newblock \emph{Physics of Fluids}, 36(9):097121.

\bibitem[{Fan et~al.(2024)Fan, Ding, Ning, Wang, Li, Yin, Chua, and Li}]{fan2024survey}
Wenqi Fan, Yujuan Ding, Liangbo Ning, Shijie Wang, Hengyun Li, Dawei Yin, Tat-Seng Chua, and Qing Li. 2024.
\newblock A survey on rag meeting llms: Towards retrieval-augmented large language models.
\newblock In \emph{Proceedings of the 30th ACM SIGKDD Conference on Knowledge Discovery and Data Mining}, pages 6491--6501.

\bibitem[{Fang et~al.(2024{\natexlab{a}})Fang, Liang, Zhang, Liu, Huang, Chen, Fan, and Chen}]{fang2023mol}
Yin Fang, Xiaozhuan Liang, Ningyu Zhang, Kangwei Liu, Rui Huang, Zhuo Chen, Xiaohui Fan, and Huajun Chen. 2024{\natexlab{a}}.
\newblock \href {https://openreview.net/pdf?id=Tlsdsb6l9n} {Mol-instructions: {A} large-scale biomolecular instruction dataset for large language models}.
\newblock In \emph{{ICLR}}. OpenReview.net.

\bibitem[{Fang et~al.(2024{\natexlab{b}})Fang, Liang, Zhang, Liu, Huang, Chen, Fan, and Chen}]{fang2024molinstructions}
Yin Fang, Xiaozhuan Liang, Ningyu Zhang, Kangwei Liu, Rui Huang, Zhuo Chen, Xiaohui Fan, and Huajun Chen. 2024{\natexlab{b}}.
\newblock \href {https://openreview.net/forum?id=Tlsdsb6l9n} {Mol-instructions: A large-scale biomolecular instruction dataset for large language models}.
\newblock In \emph{The Twelfth International Conference on Learning Representations}.

\bibitem[{Fang et~al.(2024{\natexlab{c}})Fang, Zhang, Chen, Guo, Fan, and Chen}]{fang2024domainagnostic}
Yin Fang, Ningyu Zhang, Zhuo Chen, Lingbing Guo, Xiaohui Fan, and Huajun Chen. 2024{\natexlab{c}}.
\newblock \href {https://openreview.net/forum?id=9rPyHyjfwP} {Domain-agnostic molecular generation with chemical feedback}.
\newblock In \emph{The Twelfth International Conference on Learning Representations}.

\bibitem[{Filip(2021)}]{filip2021automation}
Florin~Gheorghe Filip. 2021.
\newblock Automation and computers and their contribution to human well-being and resilience.
\newblock \emph{Studies in Informatics and Control}, 30(4):5--18.

\bibitem[{Filip(2022)}]{filip2022collaborative}
Florin~Gheorghe Filip. 2022.
\newblock Collaborative decision-making: concepts and supporting information and communication technology tools and systems.
\newblock \emph{International Journal of Computers Communications \& Control}, 17(2).

\bibitem[{Filip et~al.(2015)Filip, Ciurea, Dragomirescu, and Ivan}]{filip2015cultural}
Florin~Gheorghe Filip, Cristian Ciurea, Hora{\c{t}}iu Dragomirescu, and Ion Ivan. 2015.
\newblock Cultural heritage and modern information and communication technologies.
\newblock \emph{Technological and economic development of economy}, 21(3):441--459.

\bibitem[{Frey et~al.(2023)Frey, Soklaski, Axelrod, Samsi, G´omez-Bombarelli, Coley, and Gadepally}]{frey2023nature}
Nathan~C Frey, Ryan Soklaski, Simon Axelrod, Siddharth Samsi, Rafael G´omez-Bombarelli, Connor~W. Coley, and Vijay Gadepally. 2023.
\newblock \href {https://api.semanticscholar.org/CorpusID:262152780} {Neural scaling of deep chemical models}.
\newblock \emph{Nature Machine Intelligence}, 5:1297 -- 1305.

\bibitem[{Gao et~al.(2024{\natexlab{a}})Gao, Lan, Li, Yuan, Ding, Zhou, Xu, and Li}]{gao2024large}
Chen Gao, Xiaochong Lan, Nian Li, Yuan Yuan, Jingtao Ding, Zhilun Zhou, Fengli Xu, and Yong Li. 2024{\natexlab{a}}.
\newblock Large language models empowered agent-based modeling and simulation: A survey and perspectives.
\newblock \emph{Humanities and Social Sciences Communications}, 11(1):1--24.

\bibitem[{Gao et~al.(2024{\natexlab{b}})Gao, Fang, Huang, Giunchiglia, Noori, Schwarz, Ektefaie, Kondic, and Zitnik}]{gao2024empowering}
Shanghua Gao, Ada Fang, Yepeng Huang, Valentina Giunchiglia, Ayush Noori, Jonathan~Richard Schwarz, Yasha Ektefaie, Jovana Kondic, and Marinka Zitnik. 2024{\natexlab{b}}.
\newblock Empowering biomedical discovery with ai agents.
\newblock \emph{Cell}, 187(22):6125--6151.

\bibitem[{Garcia and Weilbach(2023)}]{garcia2023if}
Giselle~Gonzalez Garcia and Christian Weilbach. 2023.
\newblock If the sources could talk: Evaluating large language models for research assistance in history.
\newblock \emph{arXiv preprint arXiv:2310.10808}.

\bibitem[{Ghafarollahi and Buehler(2024)}]{ghafarollahi2024protagents}
Alireza Ghafarollahi and Markus~J Buehler. 2024.
\newblock Protagents: protein discovery via large language model multi-agent collaborations combining physics and machine learning.
\newblock \emph{Digital Discovery}.

\bibitem[{GLM(2024)}]{glm2024chatglm}
Team GLM. 2024.
\newblock Chatglm: A family of large language models from glm-130b to glm-4 all tools.
\newblock \emph{arXiv preprint arXiv:2406.12793}.

\bibitem[{Gou et~al.(2024)Gou, Shao, Gong, yelong shen, Yang, Huang, Duan, and Chen}]{gou2024tora}
Zhibin Gou, Zhihong Shao, Yeyun Gong, yelong shen, Yujiu Yang, Minlie Huang, Nan Duan, and Weizhu Chen. 2024.
\newblock \href {https://openreview.net/forum?id=Ep0TtjVoap} {To{RA}: A tool-integrated reasoning agent for mathematical problem solving}.
\newblock In \emph{The Twelfth International Conference on Learning Representations}.

\bibitem[{Griffin et~al.(2023)Griffin, Kleinberg, Mozes, Mai, Vau, Caldwell, and Marvor-Parker}]{griffin2023susceptibility}
Lewis~D Griffin, Bennett Kleinberg, Maximilian Mozes, Kimberly~T Mai, Maria Vau, Matthew Caldwell, and Augustine Marvor-Parker. 2023.
\newblock Susceptibility to influence of large language models.
\newblock \emph{arXiv preprint arXiv:2303.06074}.

\bibitem[{Gujral et~al.(2024)Gujral, Awaldhi, Jain, Bhandula, and Chakraborty}]{gujral2024can}
Pratik Gujral, Kshitij Awaldhi, Navya Jain, Bhavuk Bhandula, and Abhijnan Chakraborty. 2024.
\newblock Can llms help predict elections?(counter) evidence from the world's largest democracy.
\newblock \emph{arXiv preprint arXiv:2405.07828}.

\bibitem[{Guo et~al.(2023)Guo, Cheng, Wang, Li, and Liu}]{guo-etal-2023-prompt}
Zhicheng Guo, Sijie Cheng, Yile Wang, Peng Li, and Yang Liu. 2023.
\newblock \href {https://doi.org/10.18653/v1/2023.findings-acl.693} {Prompt-guided retrieval augmentation for non-knowledge-intensive tasks}.
\newblock In \emph{Findings of the Association for Computational Linguistics: ACL 2023}, pages 10896--10912, Toronto, Canada. Association for Computational Linguistics.

\bibitem[{Han et~al.(2024)Han, Wan, Chen, Yu, and Chen}]{han2024generalist}
Yang Han, Ziping Wan, Lu~Chen, Kai Yu, and Xin Chen. 2024.
\newblock From generalist to specialist: A survey of large language models for chemistry.
\newblock \emph{arXiv preprint arXiv:2412.19994}.

\bibitem[{He-Yueya et~al.(2023)He-Yueya, Poesia, Wang, and Goodman}]{he2023solving}
Joy He-Yueya, Gabriel Poesia, Rose~E Wang, and Noah~D Goodman. 2023.
\newblock Solving math word problems by combining language models with symbolic solvers.
\newblock \emph{arXiv preprint arXiv:2304.09102}.

\bibitem[{Heseltine and Clemm~von Hohenberg(2024)}]{heseltine2024large}
Michael Heseltine and Bernhard Clemm~von Hohenberg. 2024.
\newblock Large language models as a substitute for human experts in annotating political text.
\newblock \emph{Research \& Politics}, 11(1):20531680241236239.

\bibitem[{Huang et~al.(2024{\natexlab{a}})Huang, Qu, Cousins, Johnson, Yin, Shah, Zhou, Altman, Wang, and Cong}]{huang2024crispr}
Kaixuan Huang, Yuanhao Qu, Henry Cousins, William~A Johnson, Di~Yin, Mihir Shah, Denny Zhou, Russ Altman, Mengdi Wang, and Le~Cong. 2024{\natexlab{a}}.
\newblock Crispr-gpt: An llm agent for automated design of gene-editing experiments.
\newblock \emph{arXiv preprint arXiv:2404.18021}.

\bibitem[{Huang et~al.(2024{\natexlab{b}})Huang, Yu, Ma, Zhong, Feng, Wang, Chen, Peng, Feng, Qin et~al.}]{huang2024survey}
Lei Huang, Weijiang Yu, Weitao Ma, Weihong Zhong, Zhangyin Feng, Haotian Wang, Qianglong Chen, Weihua Peng, Xiaocheng Feng, Bing Qin, et~al. 2024{\natexlab{b}}.
\newblock A survey on hallucination in large language models: Principles, taxonomy, challenges, and open questions.
\newblock \emph{ACM Transactions on Information Systems}.

\bibitem[{Jaech et~al.(2024)Jaech, Kalai, Lerer, Richardson, El-Kishky, Low, Helyar, Madry, Beutel, Carney et~al.}]{jaech2024openai}
Aaron Jaech, Adam Kalai, Adam Lerer, Adam Richardson, Ahmed El-Kishky, Aiden Low, Alec Helyar, Aleksander Madry, Alex Beutel, Alex Carney, et~al. 2024.
\newblock Openai o1 system card.
\newblock \emph{arXiv preprint arXiv:2412.16720}.

\bibitem[{Jaiswal et~al.(2024)Jaiswal, Jain, Popat, Anand, Dharmadhikari, Marathe, and Shah}]{jaiswal2024improving}
Raj Jaiswal, Dhruv Jain, Harsh~Parimal Popat, Avinash Anand, Abhishek Dharmadhikari, Atharva Marathe, and Rajiv~Ratn Shah. 2024.
\newblock Improving physics reasoning in large language models using mixture of refinement agents.
\newblock \emph{arXiv preprint arXiv:2412.00821}.

\bibitem[{Jang et~al.(2022)Jang, Ye, Yang, Shin, Han, Kim, Choi, and Seo}]{jang2022towards}
Joel Jang, Seonghyeon Ye, Sohee Yang, Joongbo Shin, Janghoon Han, Gyeonghun Kim, Stanley~Jungkyu Choi, and Minjoon Seo. 2022.
\newblock Towards continual knowledge learning of language models.
\newblock In \emph{ICLR}.

\bibitem[{Jeong et~al.(2024)Jeong, Sohn, Sung, and Kang}]{jeong2024improving}
Minbyul Jeong, Jiwoong Sohn, Mujeen Sung, and Jaewoo Kang. 2024.
\newblock Improving medical reasoning through retrieval and self-reflection with retrieval-augmented large language models.
\newblock \emph{arXiv preprint arXiv:2401.15269}.

\bibitem[{Jiang et~al.(2022)Jiang, Li, Tworkowski, Czechowski, Odrzyg{\'o}{\'z}d{\'z}, Mi{\l}o{\'s}, Wu, and Jamnik}]{jiang2022thor}
Albert~Qiaochu Jiang, Wenda Li, Szymon Tworkowski, Konrad Czechowski, Tomasz Odrzyg{\'o}{\'z}d{\'z}, Piotr Mi{\l}o{\'s}, Yuhuai Wu, and Mateja Jamnik. 2022.
\newblock Thor: Wielding hammers to integrate language models and automated theorem provers.
\newblock \emph{Advances in Neural Information Processing Systems}, 35:8360--8373.

\bibitem[{Jiang et~al.(2024)Jiang, Fonseca, and Cohen}]{jiang-etal-2024-leanreasoner}
Dongwei Jiang, Marcio Fonseca, and Shay Cohen. 2024.
\newblock \href {https://doi.org/10.18653/v1/2024.naacl-long.416} {{L}ean{R}easoner: Boosting complex logical reasoning with lean}.
\newblock In \emph{Proceedings of the 2024 Conference of the North American Chapter of the Association for Computational Linguistics: Human Language Technologies (Volume 1: Long Papers)}, pages 7497--7510, Mexico City, Mexico. Association for Computational Linguistics.

\bibitem[{Jin et~al.(2024{\natexlab{a}})Jin, Xue, Wang, Kang, Ye, Zhou, Du, and Zhang}]{jin2024prollm}
Mingyu Jin, Haochen Xue, Zhenting Wang, Boming Kang, Ruosong Ye, Kaixiong Zhou, Mengnan Du, and Yongfeng Zhang. 2024{\natexlab{a}}.
\newblock \href {https://openreview.net/forum?id=2nTzomzjjb} {Pro{LLM}: Protein chain-of-thoughts enhanced {LLM} for protein-protein interaction prediction}.
\newblock In \emph{First Conference on Language Modeling}.

\bibitem[{Jin et~al.(2024{\natexlab{b}})Jin, Yang, Chen, and Lu}]{btae075_jin}
Qiao Jin, Yifan Yang, Qingyu Chen, and Zhiyong Lu. 2024{\natexlab{b}}.
\newblock \href {https://doi.org/10.1093/bioinformatics/btae075} {Genegpt: augmenting large language models with domain tools for improved access to biomedical information}.
\newblock \emph{Bioinformatics}, 40(2):btae075.

\bibitem[{Jin et~al.(2024{\natexlab{c}})Jin, Zhao, Wang, Chen, Zhu, Xiao, and Wang}]{jin2024agentreview}
Yiqiao Jin, Qinlin Zhao, Yiyang Wang, Hao Chen, Kaijie Zhu, Yijia Xiao, and Jindong Wang. 2024{\natexlab{c}}.
\newblock Agentreview: Exploring peer review dynamics with llm agents.
\newblock \emph{arXiv preprint arXiv:2406.12708}.

\bibitem[{Jumper et~al.(2021)Jumper, Evans, Pritzel, Green, Figurnov, Ronneberger, Tunyasuvunakool, Bates, {\v{Z}}{\'\i}dek, Potapenko et~al.}]{jumper2021highly}
John Jumper, Richard Evans, Alexander Pritzel, Tim Green, Michael Figurnov, Olaf Ronneberger, Kathryn Tunyasuvunakool, Russ Bates, Augustin {\v{Z}}{\'\i}dek, Anna Potapenko, et~al. 2021.
\newblock Highly accurate protein structure prediction with alphafold.
\newblock \emph{nature}, 596(7873):583--589.

\bibitem[{Kaplan et~al.(2020)Kaplan, McCandlish, Henighan, Brown, Chess, Child, Gray, Radford, Wu, and Amodei}]{kaplan2020scaling}
Jared Kaplan, Sam McCandlish, Tom Henighan, Tom~B Brown, Benjamin Chess, Rewon Child, Scott Gray, Alec Radford, Jeffrey Wu, and Dario Amodei. 2020.
\newblock Scaling laws for neural language models.
\newblock \emph{arXiv preprint arXiv:2001.08361}.

\bibitem[{Khaliq et~al.(2024)Khaliq, Chang, Ma, Pflugfelder, and Mileti{\'c}}]{khaliq2024ragar}
M~Abdul Khaliq, P~Chang, M~Ma, Bernhard Pflugfelder, and F~Mileti{\'c}. 2024.
\newblock Ragar, your falsehood radar: Rag-augmented reasoning for political fact-checking using multimodal large language models.
\newblock \emph{arXiv preprint arXiv:2404.12065}.

\bibitem[{Kim and Baek(2024)}]{kim2024exploring}
Minsang Kim and Seungjun Baek. 2024.
\newblock Exploring large language models on cross-cultural values in connection with training methodology.
\newblock \emph{arXiv preprint arXiv:2412.08846}.

\bibitem[{Koldunov and Jung(2024)}]{koldunov2024local}
Nikolay Koldunov and Thomas Jung. 2024.
\newblock Local climate services for all, courtesy of large language models.
\newblock \emph{Communications Earth \& Environment}, 5(1):13.

\bibitem[{Lample et~al.(2022)Lample, Lacroix, Lachaux, Rodriguez, Hayat, Lavril, Ebner, and Martinet}]{lample2022hypertree}
Guillaume Lample, Timothee Lacroix, Marie-Anne Lachaux, Aurelien Rodriguez, Amaury Hayat, Thibaut Lavril, Gabriel Ebner, and Xavier Martinet. 2022.
\newblock Hypertree proof search for neural theorem proving.
\newblock \emph{Advances in neural information processing systems}, 35:26337--26349.

\bibitem[{Lewkowycz et~al.(2022)Lewkowycz, Andreassen, Dohan, Dyer, Michalewski, Ramasesh, Slone, Anil, Schlag, Gutman-Solo et~al.}]{lewkowycz2022solving}
Aitor Lewkowycz, Anders Andreassen, David Dohan, Ethan Dyer, Henryk Michalewski, Vinay Ramasesh, Ambrose Slone, Cem Anil, Imanol Schlag, Theo Gutman-Solo, et~al. 2022.
\newblock Solving quantitative reasoning problems with language models.
\newblock \emph{Advances in Neural Information Processing Systems}, 35:3843--3857.

\bibitem[{Li et~al.(2024b)Li, Teney, Yang, Wen, Xie, and Wang}]{li2024culturepark}
Cheng Li, Damien Teney, Linyi Yang, Qingsong Wen, Xing Xie, and Jindong Wang. 2024b.
\newblock Culturepark: Boosting cross-cultural understanding in large language models.
\newblock \emph{arXiv preprint arXiv:2405.15145}.

\bibitem[{Li et~al.(2024{\natexlab{a}})Li, Wang, Zhang, and Zong}]{li-etal-2024-improving-context}
Chong Li, Shaonan Wang, Jiajun Zhang, and Chengqing Zong. 2024{\natexlab{a}}.
\newblock \href {https://doi.org/10.18653/v1/2024.naacl-long.445} {Improving in-context learning of multilingual generative language models with cross-lingual alignment}.
\newblock In \emph{Proceedings of the 2024 Conference of the North American Chapter of the Association for Computational Linguistics: Human Language Technologies (Volume 1: Long Papers)}, pages 8058--8076, Mexico City, Mexico. Association for Computational Linguistics.

\bibitem[{Li et~al.(2024{\natexlab{b}})Li, Yang, Zhang, Lu, Wang, and Zong}]{DBLP:conf/acl/LiYZLWZ24}
Chong Li, Wen Yang, Jiajun Zhang, Jinliang Lu, Shaonan Wang, and Chengqing Zong. 2024{\natexlab{b}}.
\newblock \href {https://doi.org/10.18653/v1/2024.findings-acl.30} {{X}-instruction: Aligning language model in low-resource languages with self-curated cross-lingual instructions}.
\newblock In \emph{Findings of the Association for Computational Linguistics: ACL 2024}, pages 546--566, Bangkok, Thailand. Association for Computational Linguistics.

\bibitem[{Li et~al.(2024{\natexlab{c}})Li, Long, Zhou, and Xiao}]{li2024screader}
Cong Li, Qingqing Long, Yuanchun Zhou, and Meng Xiao. 2024{\natexlab{c}}.
\newblock screader: Prompting large language models to interpret scrna-seq data.
\newblock \emph{arXiv preprint arXiv:2412.18156}.

\bibitem[{Li et~al.(2023{\natexlab{a}})Li, Al~Kader~Hammoud, Itani, Khizbullin, and Ghanem}]{li2023a}
Guohao Li, Hasan~Abed Al~Kader~Hammoud, Hani Itani, Dmitrii Khizbullin, and Bernard Ghanem. 2023{\natexlab{a}}.
\newblock Camel: communicative agents for "mind" exploration of large language model society.
\newblock In \emph{Proceedings of the 37th International Conference on Neural Information Processing Systems}, NIPS '23, Red Hook, NY, USA. Curran Associates Inc.

\bibitem[{Li et~al.(2024{\natexlab{d}})Li, Lai, Li, Ren, Zhang, Kang, Wang, Li, Zhang, Ma et~al.}]{li2024agent}
Junkai Li, Yunghwei Lai, Weitao Li, Jingyi Ren, Meng Zhang, Xinhui Kang, Siyu Wang, Peng Li, Ya-Qin Zhang, Weizhi Ma, et~al. 2024{\natexlab{d}}.
\newblock Agent hospital: A simulacrum of hospital with evolvable medical agents.
\newblock \emph{arXiv preprint arXiv:2405.02957}.

\bibitem[{Li et~al.(2024{\natexlab{e}})Li, Li, Chen, Gui, Yang, Yu, Wang, Cai, Zhou, Shen et~al.}]{li2024political}
Lincan Li, Jiaqi Li, Catherine Chen, Fred Gui, Hongjia Yang, Chenxiao Yu, Zhengguang Wang, Jianing Cai, Junlong~Aaron Zhou, Bolin Shen, et~al. 2024{\natexlab{e}}.
\newblock Political-llm: Large language models in political science.
\newblock \emph{arXiv preprint arXiv:2412.06864}.

\bibitem[{Li et~al.(2024{\natexlab{f}})Li, Gao, Li, Li, and Liao}]{li-etal-2024-econagent}
Nian Li, Chen Gao, Mingyu Li, Yong Li, and Qingmin Liao. 2024{\natexlab{f}}.
\newblock \href {https://doi.org/10.18653/v1/2024.acl-long.829} {{E}con{A}gent: Large language model-empowered agents for simulating macroeconomic activities}.
\newblock In \emph{Proceedings of the 62nd Annual Meeting of the Association for Computational Linguistics (Volume 1: Long Papers)}, pages 15523--15536, Bangkok, Thailand. Association for Computational Linguistics.

\bibitem[{Li et~al.(2024{\natexlab{g}})Li, Liu, Luo, Wang, He, Kawaguchi, Chua, and Tian}]{li2024towards}
Sihang Li, Zhiyuan Liu, Yanchen Luo, Xiang Wang, Xiangnan He, Kenji Kawaguchi, Tat-Seng Chua, and Qi~Tian. 2024{\natexlab{g}}.
\newblock \href {https://openreview.net/forum?id=xI4yNlkaqh} {Towards 3d molecule-text interpretation in language models}.
\newblock In \emph{The Twelfth International Conference on Learning Representations}.

\bibitem[{Li et~al.(2023{\natexlab{b}})Li, Li, Liu, Xie, Li, Wang, Li, and Zhong}]{li2023label}
Zongxi Li, Xianming Li, Yuzhang Liu, Haoran Xie, Jing Li, Fu-lee Wang, Qing Li, and Xiaoqin Zhong. 2023{\natexlab{b}}.
\newblock Label supervised llama finetuning.
\newblock \emph{arXiv preprint arXiv:2310.01208}.

\bibitem[{Liang et~al.(2024)Liang, Zhang, Ma, Zhang, Zhao, Xiang, Zong, and Zhou}]{liang2024document}
Yupu Liang, Yaping Zhang, Cong Ma, Zhiyang Zhang, Yang Zhao, Lu~Xiang, Chengqing Zong, and Yu~Zhou. 2024.
\newblock Document image machine translation with dynamic multi-pre-trained models assembling.
\newblock In \emph{Proceedings of the 2024 Conference of the North American Chapter of the Association for Computational Linguistics: Human Language Technologies (Volume 1: Long Papers)}, pages 7084--7095.

\bibitem[{Liu et~al.(2024{\natexlab{a}})Liu, Feng, Wang, Wang, Liu, Zhao, Dengr, Ruan, Dai, Guo et~al.}]{liu2024deepseek}
Aixin Liu, Bei Feng, Bin Wang, Bingxuan Wang, Bo~Liu, Chenggang Zhao, Chengqi Dengr, Chong Ruan, Damai Dai, Daya Guo, et~al. 2024{\natexlab{a}}.
\newblock Deepseek-v2: A strong, economical, and efficient mixture-of-experts language model.
\newblock \emph{arXiv preprint arXiv:2405.04434}.

\bibitem[{Liu et~al.(2024{\natexlab{b}})Liu, Feng, Xue, Wang, Wu, Lu, Zhao, Deng, Zhang, Ruan et~al.}]{liu2024deepseekv3}
Aixin Liu, Bei Feng, Bing Xue, Bingxuan Wang, Bochao Wu, Chengda Lu, Chenggang Zhao, Chengqi Deng, Chenyu Zhang, Chong Ruan, et~al. 2024{\natexlab{b}}.
\newblock Deepseek-v3 technical report.
\newblock \emph{arXiv preprint arXiv:2412.19437}.

\bibitem[{Liu et~al.(2024{\natexlab{c}})Liu, Chen, and Wang}]{liu2024genoagent}
Haoyang Liu, Shuyu Chen, and Haohan Wang. 2024{\natexlab{c}}.
\newblock \href {https://openreview.net/forum?id=v7aeTmfGOu} {Genoagent: A baseline method for {LLM}-based exploration of gene expression data in alignment with bioinformaticians}.

\bibitem[{Liu et~al.(2025)Liu, Yin, Luo, and Wang}]{liu2025integrating}
Hongxuan Liu, Haoyu Yin, Zhiyao Luo, and Xiaonan Wang. 2025.
\newblock Integrating chemistry knowledge in large language models via prompt engineering.
\newblock \emph{Synthetic and Systems Biotechnology}, 10(1):23--38.

\bibitem[{Liu et~al.(2024{\natexlab{d}})Liu, Huang, Chen, and Fang}]{liu2024drak}
Jinzhe Liu, Xiangsheng Huang, Zhuo Chen, and Yin Fang. 2024{\natexlab{d}}.
\newblock Drak: Unlocking molecular insights with domain-specific retrieval-augmented knowledge in llms.
\newblock In \emph{CCF International Conference on Natural Language Processing and Chinese Computing}, pages 255--267. Springer.

\bibitem[{Liu et~al.(2023{\natexlab{a}})Liu, Yuan, Fu, Jiang, Hayashi, and Neubig}]{liu2023pre}
Pengfei Liu, Weizhe Yuan, Jinlan Fu, Zhengbao Jiang, Hiroaki Hayashi, and Graham Neubig. 2023{\natexlab{a}}.
\newblock Pre-train, prompt, and predict: A systematic survey of prompting methods in natural language processing.
\newblock \emph{ACM Computing Surveys}, 55(9):1--35.

\bibitem[{Liu et~al.(2023{\natexlab{b}})Liu, Hu, Zhou, Ding, Li, Zeng, He, Chen, Jiang, Zhou et~al.}]{liu2023mathematical}
Wentao Liu, Hanglei Hu, Jie Zhou, Yuyang Ding, Junsong Li, Jiayi Zeng, Mengliang He, Qin Chen, Bo~Jiang, Aimin Zhou, et~al. 2023{\natexlab{b}}.
\newblock Mathematical language models: A survey.
\newblock \emph{arXiv preprint arXiv:2312.07622}.

\bibitem[{Liu et~al.(2023{\natexlab{c}})Liu, Singh, Freeman, Co-Reyes, and Liu}]{liu2023improving}
Yixin Liu, Avi Singh, C~Daniel Freeman, John~D Co-Reyes, and Peter~J Liu. 2023{\natexlab{c}}.
\newblock Improving large language model fine-tuning for solving math problems.
\newblock \emph{arXiv preprint arXiv:2310.10047}.

\bibitem[{Lu et~al.(2023)Lu, Yuan, Yuan, Lin, Lin, Tan, Zhou, and Zhou}]{lu2023instag}
Keming Lu, Hongyi Yuan, Zheng Yuan, Runji Lin, Junyang Lin, Chuanqi Tan, Chang Zhou, and Jingren Zhou. 2023.
\newblock \# instag: Instruction tagging for analyzing supervised fine-tuning of large language models.
\newblock In \emph{The Twelfth International Conference on Learning Representations}.

\bibitem[{Luo et~al.(2023{\natexlab{a}})Luo, Yang, Hong, Liu, and Nie}]{luo2023molfm}
Yizhen Luo, Kai Yang, Massimo Hong, Xing~Yi Liu, and Zaiqing Nie. 2023{\natexlab{a}}.
\newblock Molfm: A multimodal molecular foundation model.
\newblock \emph{arXiv preprint arXiv:2307.09484}.

\bibitem[{Luo et~al.(2023{\natexlab{b}})Luo, Zhang, Fan, Yang, Wu, Qiao, and Nie}]{luo2023biomedgpt}
Yizhen Luo, Jiahuan Zhang, Siqi Fan, Kai Yang, Yushuai Wu, Mu~Qiao, and Zaiqing Nie. 2023{\natexlab{b}}.
\newblock Biomedgpt: Open multimodal generative pre-trained transformer for biomedicine.
\newblock \emph{arXiv preprint arXiv:2308.09442}.

\bibitem[{Lv et~al.(2024)Lv, Lin, Li, Liu, Cui, Yu-Chian~Chen, Yuan, and Tian}]{lv2024prollama}
Liuzhenghao Lv, Zongying Lin, Hao Li, Yuyang Liu, Jiaxi Cui, Calvin Yu-Chian~Chen, Li~Yuan, and Yonghong Tian. 2024.
\newblock Prollama: A protein large language model for multi-task protein language processing.
\newblock \emph{arXiv e-prints}, pages arXiv--2402.

\bibitem[{M.~Bran et~al.(2024)M.~Bran, Cox, Schilter, Baldassari, White, and Schwaller}]{m2024augmenting}
Andres M.~Bran, Sam Cox, Oliver Schilter, Carlo Baldassari, Andrew~D White, and Philippe Schwaller. 2024.
\newblock Augmenting large language models with chemistry tools.
\newblock \emph{Nature Machine Intelligence}, pages 1--11.

\bibitem[{Ma et~al.(2024)Ma, Wang, Guo, Sun, Tenenbaum, Rus, Gan, and Matusik}]{Ma2024discovery}
Pingchuan Ma, Tsun-Hsuan Wang, Minghao Guo, Zhiqing Sun, Joshua~B. Tenenbaum, Daniela Rus, Chuang Gan, and Wojciech Matusik. 2024.
\newblock Llm and simulation as bilevel optimizers: a new paradigm to advance physical scientific discovery.
\newblock In \emph{Proceedings of the 41st International Conference on Machine Learning}, ICML'24. JMLR.org.

\bibitem[{Masoud et~al.(2023)Masoud, Liu, Ferianc, Treleaven, and Rodrigues}]{masoud2023cultural}
Reem~I Masoud, Ziquan Liu, Martin Ferianc, Philip Treleaven, and Miguel Rodrigues. 2023.
\newblock Cultural alignment in large language models: An explanatory analysis based on hofstede's cultural dimensions.
\newblock \emph{arXiv preprint arXiv:2309.12342}.

\bibitem[{McNaughton et~al.(2024)McNaughton, Sankar~Ramalaxmi, Kruel, Knutson, Varikoti, and Kumar}]{mcnaughton2024cactus}
Andrew~D McNaughton, Gautham~Krishna Sankar~Ramalaxmi, Agustin Kruel, Carter~R Knutson, Rohith~A Varikoti, and Neeraj Kumar. 2024.
\newblock Cactus: Chemistry agent connecting tool usage to science.
\newblock \emph{ACS omega}, 9(46):46563--46573.

\bibitem[{Mehta et~al.(2023)Mehta, Abbate, Wang, Rothstein, Char, Schneider, Kolemen, Rea, and Garnier}]{mehta2023towards}
Viraj Mehta, Joseph Abbate, Allen Wang, Andrew Rothstein, Ian Char, Jeff Schneider, Egemen Kolemen, Cristina Rea, and Darren Garnier. 2023.
\newblock \href {https://openreview.net/forum?id=yGVChrbJ4E} {Towards {LLM}s as operational copilots for fusion reactors}.
\newblock In \emph{NeurIPS 2023 AI for Science Workshop}.

\bibitem[{Meng et~al.(2024)Meng, Xia, and Chen}]{meng2024simpo}
Yu~Meng, Mengzhou Xia, and Danqi Chen. 2024.
\newblock Simpo: Simple preference optimization with a reference-free reward.
\newblock \emph{arXiv preprint arXiv:2405.14734}.

\bibitem[{Najafi and Varol(2024)}]{najafi2024turkishbertweet}
Ali Najafi and Onur Varol. 2024.
\newblock Turkishbertweet: Fast and reliable large language model for social media analysis.
\newblock \emph{Expert Systems with Applications}, 255:124737.

\bibitem[{Naous et~al.(2023)Naous, Ryan, Ritter, and Xu}]{naous2023having}
Tarek Naous, Michael~J Ryan, Alan Ritter, and Wei Xu. 2023.
\newblock Having beer after prayer? measuring cultural bias in large language models.
\newblock \emph{arXiv preprint arXiv:2305.14456}.

\bibitem[{Nguyen et~al.(2023)Nguyen, Ting, Ciuc{\u{a}}, O'Neill, Sun, Jab{\l}o{\'n}ska, Kruk, Perkowski, Miller, Li et~al.}]{nguyen2023astrollama}
Tuan~Dung Nguyen, Yuan-Sen Ting, Ioana Ciuc{\u{a}}, Charlie O'Neill, Ze-Chang Sun, Maja Jab{\l}o{\'n}ska, Sandor Kruk, Ernest Perkowski, Jack Miller, Jason Li, et~al. 2023.
\newblock \href {https://doi.org/10.18653/v1/2023.wiesp-1.7} {Astrollama: Towards specialized foundation models in astronomy}.
\newblock In \emph{Proceedings of the Second Workshop on Information Extraction from Scientific Publications}, pages 49--55, Bali, Indonesia. Association for Computational Linguistics.

\bibitem[{O{'}Donoghue et~al.(2023)O{'}Donoghue, Shtedritski, Ginger, Abboud, Ghareeb, and Rodriques}]{odonoghue-etal-2023-bioplanner}
Odhran O{'}Donoghue, Aleksandar Shtedritski, John Ginger, Ralph Abboud, Ali Ghareeb, and Samuel Rodriques. 2023.
\newblock \href {https://doi.org/10.18653/v1/2023.emnlp-main.162} {{B}io{P}lanner: Automatic evaluation of {LLM}s on protocol planning in biology}.
\newblock In \emph{Proceedings of the 2023 Conference on Empirical Methods in Natural Language Processing}, pages 2676--2694, Singapore. Association for Computational Linguistics.

\bibitem[{OpenAI(2022)}]{OpenAI2022}
OpenAI. 2022.
\newblock {Introducing ChatGPT}.
\newblock \url{https://openai.com/blog/chatgpt}.

\bibitem[{OpenAI(2023)}]{2023arXiv230308774OpenAI}
OpenAI. 2023.
\newblock \href {https://doi.org/10.48550/arXiv.2303.08774} {{GPT-4 Technical Report}}.
\newblock \emph{arXiv e-prints}, arXiv:2303.08774.

\bibitem[{Otieno(2024)}]{otieno2024framework}
Peter~Nyansera Otieno. 2024.
\newblock Framework for building linguistic corpora for a large language model project for the heritage nubian language of kenya.
\newblock \emph{Journal of Languages, Linguistics and Literary Studies}, 4(3):139--144.

\bibitem[{Ouyang et~al.(2022)Ouyang, Wu, Jiang, Almeida, Wainwright, Mishkin, Zhang et~al.}]{Ouyang2022}
Long Ouyang, Jeff Wu, Xu~Jiang, Diogo Almeida, Carroll~L. Wainwright, Pamela Mishkin, Chong Zhang, et~al. 2022.
\newblock Training language models to follow instructions with human feedback.
\newblock In \emph{Proceedings of the 36th International Conference on Neural Information Processing Systems}, NIPS '22, Red Hook, NY, USA. Curran Associates Inc.

\bibitem[{Pan et~al.(2024)Pan, Nguyen, Arora, Accomazzi, Ghosal, and Ting}]{pan2024astromlab}
Rui Pan, Tuan~Dung Nguyen, Hardik Arora, Alberto Accomazzi, Tirthankar Ghosal, and Yuan-Sen Ting. 2024.
\newblock Astromlab 2: Astrollama-2-70b model and benchmarking specialised llms for astronomy.
\newblock In \emph{SC24-W: Workshops of the International Conference for High Performance Computing, Networking, Storage and Analysis}, pages 87--96. IEEE.

\bibitem[{Pang et~al.(2024)Pang, Hong, Zhou, Lv, Yang, Liang, Han, and Zhang}]{pang2024physics}
Xinyu Pang, Ruixin Hong, Zhanke Zhou, Fangrui Lv, Xinwei Yang, Zhilong Liang, Bo~Han, and Changshui Zhang. 2024.
\newblock Physics reasoner: Knowledge-augmented reasoning for solving physics problems with large language models.
\newblock \emph{arXiv preprint arXiv:2412.13791}.

\bibitem[{Park et~al.(2023)Park, O'Brien, Cai, Morris, Liang, and Bernstein}]{park2023}
Joon~Sung Park, Joseph O'Brien, Carrie~Jun Cai, Meredith~Ringel Morris, Percy Liang, and Michael~S. Bernstein. 2023.
\newblock \href {https://doi.org/10.1145/3586183.3606763} {Generative agents: Interactive simulacra of human behavior}.
\newblock In \emph{Proceedings of the 36th Annual ACM Symposium on User Interface Software and Technology}, UIST '23, New York, NY, USA. Association for Computing Machinery.

\bibitem[{Patil et~al.(2023)Patil, Zhang, Wang, and Gonzalez}]{patil2023gorilla}
Shishir~G Patil, Tianjun Zhang, Xin Wang, and Joseph~E Gonzalez. 2023.
\newblock Gorilla: Large language model connected with massive apis.
\newblock \emph{arXiv preprint arXiv:2305.15334}.

\bibitem[{Perkowski et~al.(2024)Perkowski, Pan, Nguyen, Ting, Kruk, Zhang, O’Neill, Jablonska, Sun, Smith et~al.}]{perkowski2024astrollama}
Ernest Perkowski, Rui Pan, Tuan~Dung Nguyen, Yuan-Sen Ting, Sandor Kruk, Tong Zhang, Charlie O’Neill, Maja Jablonska, Zechang Sun, Michael~J Smith, et~al. 2024.
\newblock Astrollama-chat: Scaling astrollama with conversational and diverse datasets.
\newblock \emph{Research Notes of the AAS}, 8(1):7.

\bibitem[{Qin et~al.(2024)Qin, Liang, Ye, Zhu, Yan, Lu, Lin, Cong, Tang, Qian, Zhao, Hong, Tian, Xie, Zhou, Gerstein, dahai li, Liu, and Sun}]{qin2024toolllm}
Yujia Qin, Shihao Liang, Yining Ye, Kunlun Zhu, Lan Yan, Yaxi Lu, Yankai Lin, Xin Cong, Xiangru Tang, Bill Qian, Sihan Zhao, Lauren Hong, Runchu Tian, Ruobing Xie, Jie Zhou, Mark Gerstein, dahai li, Zhiyuan Liu, and Maosong Sun. 2024.
\newblock \href {https://openreview.net/forum?id=dHng2O0Jjr} {Tool{LLM}: Facilitating large language models to master 16000+ real-world {API}s}.
\newblock In \emph{The Twelfth International Conference on Learning Representations}.

\bibitem[{Radford(2018)}]{radford2018improving}
Alec Radford. 2018.
\newblock Improving language understanding by generative pre-training.

\bibitem[{Rafailov et~al.(2024)Rafailov, Sharma, Mitchell, Manning, Ermon, and Finn}]{rafailov2024direct}
Rafael Rafailov, Archit Sharma, Eric Mitchell, Christopher~D Manning, Stefano Ermon, and Chelsea Finn. 2024.
\newblock Direct preference optimization: Your language model is secretly a reward model.
\newblock \emph{Advances in Neural Information Processing Systems}, 36.

\bibitem[{Ram et~al.(2023)Ram, Levine, Dalmedigos, Muhlgay, Shashua, Leyton-Brown, and Shoham}]{ram2023context}
Ori Ram, Yoav Levine, Itay Dalmedigos, Dor Muhlgay, Amnon Shashua, Kevin Leyton-Brown, and Yoav Shoham. 2023.
\newblock In-context retrieval-augmented language models.
\newblock \emph{Transactions of the Association for Computational Linguistics}, 11:1316--1331.

\bibitem[{Ramos et~al.(2025)Ramos, Collison, and White}]{ramos2025review}
Mayk~Caldas Ramos, Christopher~J Collison, and Andrew~D White. 2025.
\newblock A review of large language models and autonomous agents in chemistry.
\newblock \emph{Chemical Science}.

\bibitem[{Ross et~al.(2022)Ross, Belgodere, Chenthamarakshan, Padhi, Mroueh, and Das}]{ross2022large}
Jerret Ross, Brian Belgodere, Vijil Chenthamarakshan, Inkit Padhi, Youssef Mroueh, and Payel Das. 2022.
\newblock Large-scale chemical language representations capture molecular structure and properties.
\newblock \emph{Nature Machine Intelligence}, 4(12):1256--1264.

\bibitem[{Rozado(2024)}]{rozado2024political}
David Rozado. 2024.
\newblock The political preferences of llms.
\newblock \emph{arXiv preprint arXiv:2402.01789}.

\bibitem[{Sanh et~al.(2022)Sanh, Webson, Raffel, Bach, Sutawika, Alyafeai, Chaffin et~al.}]{sanh2022multitask}
Victor Sanh, Albert Webson, Colin Raffel, Stephen Bach, Lintang Sutawika, Zaid Alyafeai, Antoine Chaffin, et~al. 2022.
\newblock \href {https://openreview.net/forum?id=9Vrb9D0WI4} {Multitask prompted training enables zero-shot task generalization}.
\newblock In \emph{International Conference on Learning Representations}.

\bibitem[{Schick et~al.(2023)Schick, Dwivedi-Yu, Dessi, Raileanu, Lomeli, Hambro, Zettlemoyer, Cancedda, and Scialom}]{schick2023toolformer}
Timo Schick, Jane Dwivedi-Yu, Roberto Dessi, Roberta Raileanu, Maria Lomeli, Eric Hambro, Luke Zettlemoyer, Nicola Cancedda, and Thomas Scialom. 2023.
\newblock \href {https://openreview.net/forum?id=Yacmpz84TH} {Toolformer: Language models can teach themselves to use tools}.
\newblock In \emph{Thirty-seventh Conference on Neural Information Processing Systems}.

\bibitem[{Shao et~al.(2024)Shao, Wang, Zhu, Xu, Song, Bi, Zhang, Zhang, Li, Wu et~al.}]{shao2024deepseekmath}
Zhihong Shao, Peiyi Wang, Qihao Zhu, Runxin Xu, Junxiao Song, Xiao Bi, Haowei Zhang, Mingchuan Zhang, YK~Li, Y~Wu, et~al. 2024.
\newblock Deepseekmath: Pushing the limits of mathematical reasoning in open language models.
\newblock \emph{arXiv preprint arXiv:2402.03300}.

\bibitem[{Sharma et~al.(2025)Sharma, Vaidya, Wadadekar, Bagla, Chatterjee, Hanasoge, Kumar, Mukherjee, Philip, and Singh}]{sharma2025computational}
Prateek Sharma, Bhargav Vaidya, Yogesh Wadadekar, Jasjeet Bagla, Piyali Chatterjee, Shravan Hanasoge, Prayush Kumar, Dipanjan Mukherjee, Ninan~Sajeeth Philip, and Nishant Singh. 2025.
\newblock Computational astrophysics, data science \& ai/ml in astronomy: A perspective from indian community.
\newblock \emph{arXiv preprint arXiv:2501.03876}.

\bibitem[{Tang et~al.(2024)Tang, Jin, Zhu, Yuan, Zhang, Zhou, Qu, Zhao, Tang, Zhang, Cohan, Lu, and Gerstein}]{tang2024prioritizing}
Xiangru Tang, Qiao Jin, Kunlun Zhu, Tongxin Yuan, Yichi Zhang, Wangchunshu Zhou, Meng Qu, Yilun Zhao, Jian Tang, Zhuosheng Zhang, Arman Cohan, Zhiyong Lu, and Mark Gerstein. 2024.
\newblock \href {https://openreview.net/forum?id=TBOKAvOiIy} {Prioritizing safeguarding over autonomy: Risks of {LLM} agents for science}.
\newblock In \emph{ICLR 2024 Workshop on Large Language Model (LLM) Agents}.

\bibitem[{Tinn et~al.(2023)Tinn, Cheng, Gu, Usuyama, Liu, Naumann, Gao, and Poon}]{tinn2023fine}
Robert Tinn, Hao Cheng, Yu~Gu, Naoto Usuyama, Xiaodong Liu, Tristan Naumann, Jianfeng Gao, and Hoifung Poon. 2023.
\newblock Fine-tuning large neural language models for biomedical natural language processing.
\newblock \emph{Patterns}, 4(4).

\bibitem[{{Touvron} et~al.(2023){Touvron}, {Lavril}, {Izacard}, {Martinet}, {Lachaux}, {Lacroix}, {Rozi{\`e}re}, {Goyal}, {Hambro}, {Azhar}, {Rodriguez}, {Joulin}, {Grave}, and {Lample}}]{Touvron2023}
Hugo {Touvron}, Thibaut {Lavril}, Gautier {Izacard}, Xavier {Martinet}, Marie-Anne {Lachaux}, Timoth{\'e}e {Lacroix}, Baptiste {Rozi{\`e}re}, Naman {Goyal}, Eric {Hambro}, Faisal {Azhar}, Aurelien {Rodriguez}, Armand {Joulin}, Edouard {Grave}, and Guillaume {Lample}. 2023.
\newblock \href {https://doi.org/10.48550/arXiv.2302.13971} {{LLaMA: Open and Efficient Foundation Language Models}}.
\newblock \emph{arXiv e-prints}, arXiv:2302.13971.

\bibitem[{Touvron et~al.(2023)Touvron, Martin, Stone, Albert, Almahairi, Babaei, Bashlykov, Batra, Bhargava, Bhosale et~al.}]{touvron2023llama}
Hugo Touvron, Louis Martin, Kevin Stone, Peter Albert, Amjad Almahairi, Yasmine Babaei, Nikolay Bashlykov, Soumya Batra, Prajjwal Bhargava, Shruti Bhosale, et~al. 2023.
\newblock Llama 2: Open foundation and fine-tuned chat models.
\newblock \emph{arXiv preprint arXiv:2307.09288}.

\bibitem[{Trichopoulos(2023)}]{trichopoulos2023large}
Georgios Trichopoulos. 2023.
\newblock Large language models for cultural heritage.
\newblock In \emph{Proceedings of the 2nd International Conference of the ACM Greek SIGCHI Chapter}, pages 1--5.

\bibitem[{Upadhyay et~al.(2023)Upadhyay, Ginsberg, and Callison-Burch}]{upadhyay2023improving}
Shriyash Upadhyay, Etan Ginsberg, and Chris Callison-Burch. 2023.
\newblock Improving mathematics tutoring with a code scratchpad.
\newblock In \emph{Proceedings of the 18th Workshop on Innovative Use of NLP for Building Educational Applications (BEA 2023)}, pages 20--28.

\bibitem[{Vaghefi et~al.(2023)Vaghefi, Stammbach, Muccione, Bingler, Ni, Kraus, Allen, Colesanti-Senni, Wekhof, Schimanski et~al.}]{vaghefi2023chatclimate}
Saeid~Ashraf Vaghefi, Dominik Stammbach, Veruska Muccione, Julia Bingler, Jingwei Ni, Mathias Kraus, Simon Allen, Chiara Colesanti-Senni, Tobias Wekhof, Tobias Schimanski, et~al. 2023.
\newblock Chatclimate: Grounding conversational ai in climate science.
\newblock \emph{Communications Earth \& Environment}, 4(1):480.

\bibitem[{Vaswani(2017)}]{vaswani2017attention}
A~Vaswani. 2017.
\newblock Attention is all you need.
\newblock \emph{Advances in Neural Information Processing Systems}.

\bibitem[{Wang et~al.(2024{\natexlab{a}})Wang, Long, Xiao, Cai, Wu, Meng, Wang, and Zhou}]{wang2024biorag}
Chengrui Wang, Qingqing Long, Meng Xiao, Xunxin Cai, Chengjun Wu, Zhen Meng, Xuezhi Wang, and Yuanchun Zhou. 2024{\natexlab{a}}.
\newblock Biorag: A rag-llm framework for biological question reasoning.
\newblock \emph{arXiv preprint arXiv:2408.01107}.

\bibitem[{Wang et~al.(2024{\natexlab{b}})Wang, Hu, Zhang, Chen, Du, Mao, Wang, Li, Wu, Yang et~al.}]{wang2024starwhisper}
Cunshi Wang, Xinjie Hu, Yu~Zhang, Xunhao Chen, Pengliang Du, Yiming Mao, Rui Wang, Yuyang Li, Ying Wu, Hang Yang, et~al. 2024{\natexlab{b}}.
\newblock Starwhisper telescope: Agent-based observation assistant system to approach ai astrophysicist.
\newblock \emph{arXiv preprint arXiv:2412.06412}.

\bibitem[{Wang et~al.(2023{\natexlab{a}})Wang, Xin, Zheng, Li, Liu, Cao, Huang, Xiong, Shi, Xie et~al.}]{wang2023lego}
Haiming Wang, Huajian Xin, Chuanyang Zheng, Lin Li, Zhengying Liu, Qingxing Cao, Yinya Huang, Jing Xiong, Han Shi, Enze Xie, et~al. 2023{\natexlab{a}}.
\newblock Lego-prover: Neural theorem proving with growing libraries.
\newblock \emph{arXiv preprint arXiv:2310.00656}.

\bibitem[{Wang et~al.(2024{\natexlab{c}})Wang, Ma, Feng, Zhang, Yang, Zhang, Chen, Tang, Chen, Lin et~al.}]{wang2024survey}
Lei Wang, Chen Ma, Xueyang Feng, Zeyu Zhang, Hao Yang, Jingsen Zhang, Zhiyuan Chen, Jiakai Tang, Xu~Chen, Yankai Lin, et~al. 2024{\natexlab{c}}.
\newblock A survey on large language model based autonomous agents.
\newblock \emph{Frontiers of Computer Science}, 18(6):186345.

\bibitem[{Wang et~al.(2023{\natexlab{b}})Wang, Wei, Schuurmans, Le, Chi, Narang, Chowdhery, and Zhou}]{wang2023selfconsistency}
Xuezhi Wang, Jason Wei, Dale Schuurmans, Quoc~V Le, Ed~H. Chi, Sharan Narang, Aakanksha Chowdhery, and Denny Zhou. 2023{\natexlab{b}}.
\newblock \href {https://openreview.net/forum?id=1PL1NIMMrw} {Self-consistency improves chain of thought reasoning in language models}.
\newblock In \emph{The Eleventh International Conference on Learning Representations}.

\bibitem[{Wang and Luo(2023)}]{wang2023enhance}
Yajing Wang and Zongwei Luo. 2023.
\newblock Enhance multi-domain sentiment analysis of review texts through prompting strategies.
\newblock In \emph{2023 International Conference on High Performance Big Data and Intelligent Systems (HDIS)}, pages 1--7. IEEE.

\bibitem[{Wang et~al.(2023{\natexlab{c}})Wang, Li, Sun, and Liu}]{wang-etal-2023-self-knowledge}
Yile Wang, Peng Li, Maosong Sun, and Yang Liu. 2023{\natexlab{c}}.
\newblock \href {https://doi.org/10.18653/v1/2023.findings-emnlp.691} {Self-knowledge guided retrieval augmentation for large language models}.
\newblock In \emph{Findings of the Association for Computational Linguistics: EMNLP 2023}, pages 10303--10315, Singapore. Association for Computational Linguistics.

\bibitem[{Wang et~al.(2023{\natexlab{d}})Wang, Kordi, Mishra, Liu, Smith, Khashabi, and Hajishirzi}]{wang-etal-2023-self-instruct}
Yizhong Wang, Yeganeh Kordi, Swaroop Mishra, Alisa Liu, Noah~A. Smith, Daniel Khashabi, and Hannaneh Hajishirzi. 2023{\natexlab{d}}.
\newblock \href {https://doi.org/10.18653/v1/2023.acl-long.754} {Self-instruct: Aligning language models with self-generated instructions}.
\newblock In \emph{Proceedings of the 61st Annual Meeting of the Association for Computational Linguistics (Volume 1: Long Papers)}, pages 13484--13508, Toronto, Canada. Association for Computational Linguistics.

\bibitem[{Wang et~al.(2024{\natexlab{d}})Wang, Zhang, Momtaz, Moradi, Rastegarnia, Sahakyan, Shakeri, and Li}]{wang2024can}
Yu~Wang, Shu-Rui Zhang, Aidin Momtaz, Rahim Moradi, Fatemeh Rastegarnia, Narek Sahakyan, Soroush Shakeri, and Liang Li. 2024{\natexlab{d}}.
\newblock Can ai understand our universe? test of fine-tuning gpt by astrophysical data.
\newblock \emph{arXiv preprint arXiv:2404.10019}.

\bibitem[{Wang et~al.(2024{\natexlab{e}})Wang, Xu, Wang, Zhou, and Zhou}]{wang2024intelligent}
Zhenyu Wang, Yi~Xu, Dequan Wang, Lingfeng Zhou, and Yiqi Zhou. 2024{\natexlab{e}}.
\newblock Intelligent computing social modeling and methodological innovations in political science in the era of large language models.
\newblock \emph{arXiv preprint arXiv:2410.16301}.

\bibitem[{Wei et~al.(2022)Wei, Wang, Schuurmans, Bosma, Xia, Chi, Le, Zhou et~al.}]{wei2022chain}
Jason Wei, Xuezhi Wang, Dale Schuurmans, Maarten Bosma, Fei Xia, Ed~Chi, Quoc~V Le, Denny Zhou, et~al. 2022.
\newblock Chain-of-thought prompting elicits reasoning in large language models.
\newblock \emph{Advances in neural information processing systems}, 35:24824--24837.

\bibitem[{Wu et~al.(2024)Wu, Lin, Zhang, Zhang, Xie, and Wang}]{wu2024pmc}
Chaoyi Wu, Weixiong Lin, Xiaoman Zhang, Ya~Zhang, Weidi Xie, and Yanfeng Wang. 2024.
\newblock Pmc-llama: toward building open-source language models for medicine.
\newblock \emph{Journal of the American Medical Informatics Association}, page ocae045.

\bibitem[{Xiao et~al.(2024)Xiao, Liu, Zheng, Xie, Hao, Li, Wang, Ni, Li, Luo et~al.}]{xiao2024cellagent}
Yihang Xiao, Jinyi Liu, Yan Zheng, Xiaohan Xie, Jianye Hao, Mingzhi Li, Ruitao Wang, Fei Ni, Yuxiao Li, Jintian Luo, et~al. 2024.
\newblock Cellagent: An llm-driven multi-agent framework for automated single-cell data analysis.
\newblock \emph{bioRxiv}, pages 2024--05.

\bibitem[{Xin et~al.(2024)Xin, Guo, Shao, Ren, Zhu, Liu, Ruan, Li, and Liang}]{xin2024deepseek}
Huajian Xin, Daya Guo, Zhihong Shao, Zhizhou Ren, Qihao Zhu, Bo~Liu, Chong Ruan, Wenda Li, and Xiaodan Liang. 2024.
\newblock Deepseek-prover: Advancing theorem proving in llms through large-scale synthetic data.
\newblock \emph{arXiv preprint arXiv:2405.14333}.

\bibitem[{Xu et~al.(2024{\natexlab{a}})Xu, Sun, Zheng, Geng, Zhao, Feng, Tao, Lin, and Jiang}]{xu2024wizardlm}
Can Xu, Qingfeng Sun, Kai Zheng, Xiubo Geng, Pu~Zhao, Jiazhan Feng, Chongyang Tao, Qingwei Lin, and Daxin Jiang. 2024{\natexlab{a}}.
\newblock Wizardlm: Empowering large pre-trained language models to follow complex instructions.
\newblock In \emph{The Twelfth International Conference on Learning Representations}.

\bibitem[{Xu et~al.(2024{\natexlab{b}})Xu, Fei, Pan, Liu, Lee, and Hsu}]{xu2024faithful}
Jundong Xu, Hao Fei, Liangming Pan, Qian Liu, Mong-Li Lee, and Wynne Hsu. 2024{\natexlab{b}}.
\newblock Faithful logical reasoning via symbolic chain-of-thought.
\newblock \emph{arXiv preprint arXiv:2405.18357}.

\bibitem[{{Xu} et~al.(2025){Xu}, {Xu}, {Xiao}, {Chen}, {Yan}, {Zhang}, {Diao}, {Yang}, and {Wang}}]{xu2025ugphysics}
Xin {Xu}, Qiyun {Xu}, Tong {Xiao}, Tianhao {Chen}, Yuchen {Yan}, Jiaxin {Zhang}, Shizhe {Diao}, Can {Yang}, and Yang {Wang}. 2025.
\newblock \href {https://doi.org/10.48550/arXiv.2502.00334} {{UGPhysics: A Comprehensive Benchmark for Undergraduate Physics Reasoning with Large Language Models}}.
\newblock \emph{arXiv e-prints}, arXiv:2502.00334.

\bibitem[{Yang et~al.(2024{\natexlab{a}})Yang, Yang, Zhang, Hui, Zheng, Yu, Li, Liu, Huang, Wei et~al.}]{yang2024qwen2}
An~Yang, Baosong Yang, Beichen Zhang, Binyuan Hui, Bo~Zheng, Bowen Yu, Chengyuan Li, Dayiheng Liu, Fei Huang, Haoran Wei, et~al. 2024{\natexlab{a}}.
\newblock Qwen2. 5 technical report.
\newblock \emph{arXiv preprint arXiv:2412.15115}.

\bibitem[{Yang et~al.(2024{\natexlab{b}})Yang, Zhang, Hui, Gao, Yu, Li, Liu, Tu, Zhou, Lin et~al.}]{yang2024qwen2_math}
An~Yang, Beichen Zhang, Binyuan Hui, Bofei Gao, Bowen Yu, Chengpeng Li, Dayiheng Liu, Jianhong Tu, Jingren Zhou, Junyang Lin, et~al. 2024{\natexlab{b}}.
\newblock Qwen2. 5-math technical report: Toward mathematical expert model via self-improvement.
\newblock \emph{arXiv preprint arXiv:2409.12122}.

\bibitem[{Yang et~al.(2023)Yang, Swope, Gu, Chalamala, Song, Yu, Godil, Prenger, and Anandkumar}]{yang2023leandojo}
Kaiyu Yang, Aidan~M Swope, Alex Gu, Rahul Chalamala, Peiyang Song, Shixing Yu, Saad Godil, Ryan Prenger, and Anima Anandkumar. 2023.
\newblock \href {https://openreview.net/forum?id=g7OX2sOJtn} {Leandojo: Theorem proving with retrieval-augmented language models}.
\newblock In \emph{Thirty-seventh Conference on Neural Information Processing Systems Datasets and Benchmarks Track}.

\bibitem[{Yao et~al.(2024)Yao, Yu, Zhao, Shafran, Griffiths, Cao, and Narasimhan}]{yao2024tree}
Shunyu Yao, Dian Yu, Jeffrey Zhao, Izhak Shafran, Tom Griffiths, Yuan Cao, and Karthik Narasimhan. 2024.
\newblock Tree of thoughts: Deliberate problem solving with large language models.
\newblock \emph{Advances in Neural Information Processing Systems}, 36.

\bibitem[{Yao et~al.(2023)Yao, Zhao, Yu, Du, Shafran, Narasimhan, and Cao}]{yao2023react}
Shunyu Yao, Jeffrey Zhao, Dian Yu, Nan Du, Izhak Shafran, Karthik~R Narasimhan, and Yuan Cao. 2023.
\newblock \href {https://openreview.net/forum?id=WE_vluYUL-X} {React: Synergizing reasoning and acting in language models}.
\newblock In \emph{The Eleventh International Conference on Learning Representations}.

\bibitem[{Ye et~al.(2025)Ye, Xiang, Zhang, and Zong}]{ye-etal-2025-sweetiechat}
Jing Ye, Lu~Xiang, Yaping Zhang, and Chengqing Zong. 2025.
\newblock \href {https://aclanthology.org/2025.coling-main.312/} {{S}weetie{C}hat: A strategy-enhanced role-playing framework for diverse scenarios handling emotional support agent}.
\newblock In \emph{Proceedings of the 31st International Conference on Computational Linguistics}, pages 4646--4669, Abu Dhabi, UAE. Association for Computational Linguistics.

\bibitem[{Yoshikawa et~al.(2023)Yoshikawa, Skreta, Darvish, Arellano-Rubach, Ji, Bj{\o}rn~Kristensen, Li, Zhao, Xu, Kuramshin et~al.}]{yoshikawa2023large}
Naruki Yoshikawa, Marta Skreta, Kourosh Darvish, Sebastian Arellano-Rubach, Zhi Ji, Lasse Bj{\o}rn~Kristensen, Andrew~Zou Li, Yuchi Zhao, Haoping Xu, Artur Kuramshin, et~al. 2023.
\newblock Large language models for chemistry robotics.
\newblock \emph{Autonomous Robots}, 47(8):1057--1086.

\bibitem[{Zeng et~al.(2023)Zeng, Liu, Du, Wang, Lai, Ding, Yang, Xu, Zheng, Xia, Tam, Ma, Xue, Zhai, Chen, Liu, Zhang, Dong, and Tang}]{zeng2023glmb}
Aohan Zeng, Xiao Liu, Zhengxiao Du, Zihan Wang, Hanyu Lai, Ming Ding, Zhuoyi Yang, Yifan Xu, Wendi Zheng, Xiao Xia, Weng~Lam Tam, Zixuan Ma, Yufei Xue, Jidong Zhai, Wenguang Chen, Zhiyuan Liu, Peng Zhang, Yuxiao Dong, and Jie Tang. 2023.
\newblock \href {https://openreview.net/forum?id=-Aw0rrrPUF} {{GLM}-130b: An open bilingual pre-trained model}.
\newblock In \emph{The Eleventh International Conference on Learning Representations}.

\bibitem[{Zeng(2024)}]{zeng2024histolens}
Yifan Zeng. 2024.
\newblock Histolens: An llm-powered framework for multi-layered analysis of historical texts--a case application of yantie lun.
\newblock \emph{arXiv preprint arXiv:2411.09978}.

\bibitem[{Zhang et~al.(2023{\natexlab{a}})Zhang, Yang, Zhou, Ali~Babar, and Liu}]{zhang2023enhancing}
Boyu Zhang, Hongyang Yang, Tianyu Zhou, Muhammad Ali~Babar, and Xiao-Yang Liu. 2023{\natexlab{a}}.
\newblock Enhancing financial sentiment analysis via retrieval augmented large language models.
\newblock In \emph{Proceedings of the fourth ACM international conference on AI in finance}, pages 349--356.

\bibitem[{Zhang et~al.(2024{\natexlab{a}})Zhang, Hu, Zhoubian, Du, Yang, Wang, Yue, Dong, and Tang}]{zhang2024sciglm}
Dan Zhang, Ziniu Hu, Sining Zhoubian, Zhengxiao Du, Kaiyu Yang, Zihan Wang, Yisong Yue, Yuxiao Dong, and Jie Tang. 2024{\natexlab{a}}.
\newblock Sciglm: Training scientific language models with self-reflective instruction annotation and tuning.
\newblock \emph{arXiv preprint arXiv:2401.07950}.

\bibitem[{Zhang et~al.(2024{\natexlab{b}})Zhang, Deng, Liu, Pan, and Bing}]{zhang-etal-2024-sentiment}
Wenxuan Zhang, Yue Deng, Bing Liu, Sinno Pan, and Lidong Bing. 2024{\natexlab{b}}.
\newblock \href {https://doi.org/10.18653/v1/2024.findings-naacl.246} {Sentiment analysis in the era of large language models: A reality check}.
\newblock In \emph{Findings of the Association for Computational Linguistics: NAACL 2024}, pages 3881--3906, Mexico City, Mexico. Association for Computational Linguistics.

\bibitem[{Zhang and Gao(2023)}]{zhang2023towards}
Xuan Zhang and Wei Gao. 2023.
\newblock Towards llm-based fact verification on news claims with a hierarchical step-by-step prompting method.
\newblock \emph{arXiv preprint arXiv:2310.00305}.

\bibitem[{Zhang et~al.(2024{\natexlab{c}})Zhang, Zhang-Li, Yu, Gong, Zhou, Liu, Hou, and Li}]{zhang2024simulating}
Zheyuan Zhang, Daniel Zhang-Li, Jifan Yu, Linlu Gong, Jinchang Zhou, Zhiyuan Liu, Lei Hou, and Juanzi Li. 2024{\natexlab{c}}.
\newblock Simulating classroom education with llm-empowered agents.
\newblock \emph{arXiv preprint arXiv:2406.19226}.

\bibitem[{Zhang et~al.(2025{\natexlab{a}})Zhang, Zhang, Liang, Ma, Xiang, Zhao, Zhou, and Zong}]{zhang2025understand}
Zhiyang Zhang, Yaping Zhang, Yupu Liang, Cong Ma, Lu~Xiang, Yang Zhao, Yu~Zhou, and Chengqing Zong. 2025{\natexlab{a}}.
\newblock Understand layout and translate text: Unified feature-conductive end-to-end document image translation.
\newblock \emph{IEEE Transactions on Pattern Analysis and Machine Intelligence}.

\bibitem[{Zhang et~al.(2025{\natexlab{b}})Zhang, Zhang, Liang, Xiang, Zhao, Zhou, and Zong}]{zhang-etal-2025-chaotic}
Zhiyang Zhang, Yaping Zhang, Yupu Liang, Lu~Xiang, Yang Zhao, Yu~Zhou, and Chengqing Zong. 2025{\natexlab{b}}.
\newblock \href {https://aclanthology.org/2025.coling-main.723/} {From chaotic {OCR} words to coherent document: A fine-to-coarse zoom-out network for complex-layout document image translation}.
\newblock In \emph{Proceedings of the 31st International Conference on Computational Linguistics}, pages 10877--10890, Abu Dhabi, UAE. Association for Computational Linguistics.

\bibitem[{Zhang et~al.(2023{\natexlab{b}})Zhang, Xu, Jamasb, Chenthamarakshan, Lozano, Das, and Tang}]{zhang2023protein}
Zuobai Zhang, Minghao Xu, Arian~Rokkum Jamasb, Vijil Chenthamarakshan, Aurelie Lozano, Payel Das, and Jian Tang. 2023{\natexlab{b}}.
\newblock \href {https://openreview.net/forum?id=to3qCB3tOh9} {Protein representation learning by geometric structure pretraining}.
\newblock In \emph{The Eleventh International Conference on Learning Representations}.

\bibitem[{Zhao et~al.(2023{\natexlab{a}})Zhao, Wang, Zhang, Jin, Zhu, Chen, and Xie}]{zhao2023competeai}
Qinlin Zhao, Jindong Wang, Yixuan Zhang, Yiqiao Jin, Kaijie Zhu, Hao Chen, and Xing Xie. 2023{\natexlab{a}}.
\newblock Competeai: Understanding the competition behaviors in large language model-based agents.
\newblock \emph{arXiv preprint arXiv:2310.17512}.

\bibitem[{Zhao et~al.(2023{\natexlab{b}})Zhao, Zhou, Li, Tang, Wang, Hou, Min, Zhang, Zhang, Dong et~al.}]{zhao2023survey}
Wayne~Xin Zhao, Kun Zhou, Junyi Li, Tianyi Tang, Xiaolei Wang, Yupeng Hou, Yingqian Min, Beichen Zhang, Junjie Zhang, Zican Dong, et~al. 2023{\natexlab{b}}.
\newblock A survey of large language models.
\newblock \emph{arXiv preprint arXiv:2303.18223}.

\bibitem[{Zhao et~al.(2023{\natexlab{c}})Zhao, Zhang, and Zong}]{zhao2023transformer}
Yang Zhao, Jiajun Zhang, and Chengqing Zong. 2023{\natexlab{c}}.
\newblock Transformer: A general framework from machine translation to others.
\newblock \emph{Machine Intelligence Research}, 20(4):514--538.

\bibitem[{Zhao et~al.(2024)Zhao, Ma, Chen, Sun, Li, Xia, Xu, Zhu, Zhu, Fan, Shen, Yu, and Chen}]{zhao2024chemdfm}
Zihan Zhao, Da~Ma, Lu~Chen, Liangtai Sun, Zihao Li, Yi~Xia, Hongshen Xu, Zichen Zhu, Su~Zhu, Shuai Fan, Guodong Shen, Kai Yu, and Xin Chen. 2024.
\newblock \href {https://openreview.net/forum?id=emPxd99kTC} {Chem{DFM}: A large language foundation model for chemistry}.
\newblock In \emph{Neurips 2024 Workshop Foundation Models for Science: Progress, Opportunities, and Challenges}.

\bibitem[{Zheng et~al.(2023)Zheng, Zhang, Borgs, Chayes, and Yaghi}]{zheng2023chatgpt}
Zhiling Zheng, Oufan Zhang, Christian Borgs, Jennifer~T Chayes, and Omar~M Yaghi. 2023.
\newblock Chatgpt chemistry assistant for text mining and the prediction of mof synthesis.
\newblock \emph{Journal of the American Chemical Society}, 145(32):18048--18062.

\bibitem[{Zholus et~al.(2024)Zholus, Kuznetsov, Schutski, Shayakhmetov, Polykovskiy, Chandar, and Zhavoronkov}]{zholus2024bindgpt}
Artem Zholus, Maksim Kuznetsov, Roman Schutski, Rim Shayakhmetov, Daniil Polykovskiy, Sarath Chandar, and Alex Zhavoronkov. 2024.
\newblock Bindgpt: A scalable framework for 3d molecular design via language modeling and reinforcement learning.
\newblock \emph{arXiv preprint arXiv:2406.03686}.

\bibitem[{Zhuo et~al.(2024)Zhuo, Chi, Xu, Huang, Zheng, He, Mao, and Zhang}]{zhuo2024protllm}
Le~Zhuo, Zewen Chi, Minghao Xu, Heyan Huang, Heqi Zheng, Conghui He, Xian-Ling Mao, and Wentao Zhang. 2024.
\newblock Protllm: An interleaved protein-language llm with protein-as-word pre-training.
\newblock \emph{arXiv preprint arXiv:2403.07920}.

\bibitem[{Ziems et~al.(2024)Ziems, Held, Shaikh, Chen, Zhang, and Yang}]{10.1162/coli_a_00502}
Caleb Ziems, William Held, Omar Shaikh, Jiaao Chen, Zhehao Zhang, and Diyi Yang. 2024.
\newblock \href {https://doi.org/10.1162/coli_a_00502} {Can large language models transform computational social science?}
\newblock \emph{Computational Linguistics}, 50(1):237--291.

\bibitem[{Zong et~al.(2021)Zong, Xia, and Zhang}]{zong2021text}
Chengqing Zong, Rui Xia, and Jiajun Zhang. 2021.
\newblock \emph{Text data mining}.
\newblock Tsinghua University Press and Springer.

\bibitem[{Zvyagin et~al.(2023)Zvyagin, Brace, Hippe, Deng, Zhang, Bohorquez, Clyde, Kale, Perez-Rivera, Ma et~al.}]{zvyagin2023genslms}
Maxim Zvyagin, Alexander Brace, Kyle Hippe, Yuntian Deng, Bin Zhang, Cindy~Orozco Bohorquez, Austin Clyde, Bharat Kale, Danilo Perez-Rivera, Heng Ma, et~al. 2023.
\newblock Genslms: Genome-scale language models reveal sars-cov-2 evolutionary dynamics.
\newblock \emph{The International Journal of High Performance Computing Applications}, 37(6):683--705.

\end{thebibliography}
